\documentclass[journal]{IEEEtran}
\usepackage{placeins}
\usepackage{dblfloatfix}
\usepackage{cuted}
\usepackage{capt-of}

\usepackage{amsmath,amssymb,amsfonts}
\usepackage{bm}
\usepackage{graphicx}
\usepackage{booktabs}
\usepackage{makecell}
\usepackage{cite}
\usepackage{url}

\newtheorem{theorem}{Theorem}
\newtheorem{proposition}{Proposition}
\newtheorem{remark}{Remark}
\newtheorem{conclusionx}{Conclusion}

\newcommand{\dt}{^{t_0,t_1}}
\newcommand{\skewm}[1]{\left[#1\right]_{\times}}

\graphicspath{{figs/}}

\begin{document}

\title{Differential 6-DOF Pose Estimation with Provable First-Order Immunity to Camera Calibration Errors}

\author{Yueqiang~Zhang, Liang~Deng, Yi~Zhang, Baoqiong~Wang, Wenjun~Chen, Shuixin~Pan, Yulan~Guo, and~Qifeng~Yu%
\thanks{This work was supported in part by the National Natural Science Foundation of China under Grants 12372184, 12002215, and 62505196, and in part by the Research Team Cultivation Program of Shenzhen University under Grant 2023JCT003. (\textit{Corresponding author:
Shuixin Pan}.)}
\thanks{Yueqiang~Zhang, Liang~Deng, Baoqiong~Wang, Wenjun~Chen, Shuixin~Pan, and~Qifeng~Yu are with the State Key Laboratory of Radio Frequency Heterogeneous Integration, the Key Laboratory of Optoelectronic Devices and Systems of Ministry of Education and Guangdong Province, the Shenzhen Key Laboratory of Intelligent Optical Measurement and Detection, and the College of Physics and Optoelectronic Engineering, Shenzhen University, Shenzhen 518060, China (e-mail: yueqiang.zhang@szu.edu.cn; shuixinpan@szu.edu.cn).

Yi~Zhang and Yulan~Guo are with the School of Electronics and Communication Engineering, Sun Yat-sen University, Shenzhen 510275, China (e-mail: guoyulan@sysu.edu.cn).
}
\thanks{This work has been submitted to the IEEE for possible publication. Copyright may be transferred without notice, after which this version may no longer be accessible.}
}
\markboth{Preprint submitted to IEEE Transactions on Pattern Analysis and Machine Intelligence}%
{Differential 6-DOF Pose Estimation with Provable First-Order Immunity to Camera Calibration Errors}

\maketitle

\begin{abstract}
Accurate six-degree-of-freedom (6-DOF) motion estimation of mobile platforms is essential for robotic manipulation, autonomous systems, and structural displacement monitoring. Conventional 3D-2D methods independently estimate the absolute camera pose at each epoch and recover the platform motion through camera-to-platform extrinsic transformations, making the result susceptible to extrinsic calibration errors, especially under micro-motion. This paper presents a differential pose estimation method that directly recovers platform motion from inter-frame image displacements and known 3D control points. By differencing the perspective projection equations, adopting a depth-invariance approximation, and representing the inter-frame motion on $\mathfrak{se}(3)$, the proposed method avoids independent absolute-pose estimation and supports both monocular and multi-camera configurations. We prove that the resulting differential model is exactly immune to translational extrinsic calibration errors, while the effect of rotational extrinsic errors is bounded by the calibration error, motion magnitude, and observation geometry. We further establish the generic observability condition, derive the Cram\'er--Rao lower bound and a bias-eliminated consistent estimator, and characterize the validity boundaries of the depth-invariance and first-order approximations. Extensive synthetic and real-world experiments establish a new state of the art for 6-DOF platform micro-motion estimation, outperforming representative PnP and generalized-PnP methods in accuracy, calibration robustness, and computational efficiency. With five control points and 0.5-pixel image noise, the monocular solver achieves a combined pitch--yaw RMSE of $10.09''$, a translation RMSE of 3.70~mm, and a runtime of 0.34~ms, while the binocular solver achieves a rotation RMSE of $10.58''$, a translation RMSE of 3.91~mm, and a runtime of 0.27~ms. Code and supplementary material will be released upon publication at \url{https://github.com/zyoungszu/pami2026}.
\end{abstract}


\IEEEpeerreviewmaketitle

\section{Introduction}
\IEEEPARstart{M}{otion} estimation of mobile platforms is a fundamental capability in visual SLAM~\cite{r1,r2,r3,r4}, robotic manipulation~\cite{r5,r6,r7}, and structural displacement measurement~\cite{r8,r9,r10}. Among the various approaches, vision-based methods have been widely adopted for their non-contact nature and high precision. In a typical setup, a camera is rigidly mounted on the platform, and the extrinsic transformation between the camera and platform coordinate frames is calibrated in advance. During operation, a camera pose estimation algorithm recovers the camera motion from captured images, which is then
converted to the platform motion via the calibrated extrinsics. The accuracy of the entire pipeline thus hinges on the camera pose estimation step. The tiny-displacement regime that concerns us here is not a corner case of this pipeline but a recurring problem class of its own: robot repeatability qualification (ISO 9283), machine-tool and stage drift, loop-closure and relocalization verification, on-orbit thermal deformation, and structural monitoring all reduce to estimating a platform motion far smaller than the scene depth---precisely the regime in which calibration errors overwhelm the signal being measured. What follows is a general treatment of this class, not a single application.

Camera pose estimation methods can be broadly classified into two categories. Methods based on 2D-2D point correspondences estimate the relative camera motion from feature matches between two consecutive frames~\cite{r11,r12}, but can only recover 5-DOF poses due to the inherent scale ambiguity in monocular geometry, and additional sensors or scene priors are required to recover the absolute scale. In contrast, methods based on 3D-2D correspondences, known as the Perspective-$n$-Point (PnP) problem, utilize cooperative markers with known 3D coordinates to establish correspondences and solve for the full 6-DOF camera pose at each frame independently~\cite{r4,r9,r13,r14,r15}. Since this work targets 6-DOF motion estimation with absolute scale, we focus on the latter category.

However, regardless of which category is employed, the estimated camera pose must ultimately be transformed into the platform frame through the camera-platform extrinsics. Both approaches are therefore inevitably affected by calibration errors in these extrinsic parameters. Such errors directly degrade the accuracy of platform motion estimation, and the degradation becomes particularly severe under tiny platform displacements, where the magnitude of the calibration error can far exceed the true displacement and thereby dominate the measurement result.

To overcome this limitation, we formulate platform-motion estimation as a differential 3D-2D geometry problem rather than as the difference between
two independently estimated absolute poses. For each known 3D control point, the perspective projection equations at two consecutive epochs are subtracted to construct a direct relationship between the inter-frame image displacement and the 6-DOF platform motion. Under the depth-invariance approximation valid for small inter-frame motion, and using a first-order representation of the motion on $\mathfrak{se}(3)$, the resulting constraint is linear in the platform-motion parameters. Consequently, the platform motion can be recovered directly from the image-coordinate differences without first estimating the absolute camera pose at either epoch. The differential formulation also reveals that translational camera-to-platform extrinsic errors cancel identically, whereas rotational extrinsic errors produce a bounded perturbation coupled to the platform-motion magnitude. We refer to this property as first-order immunity to extrinsic calibration errors. The same formulation extends from a single camera to an arbitrary rigid multi-camera system by stacking the constraints contributed by the individual cameras.

The main contributions of this work are summarized as follows:
\begin{enumerate}
  \item We propose a differential 6-DOF platform-motion estimator that
  directly uses inter-frame image displacements and known 3D points,
  avoiding independent absolute-pose estimation. We derive an efficient
  closed-form linear solution for both the minimal three-point and
  overdetermined configurations. The formulation extends naturally from
  monocular to arbitrary rigid multi-camera systems.
  \item We prove that the proposed model is exactly immune to translational
  extrinsic calibration errors, while the effect of rotational extrinsic
  errors is bounded by the calibration error, motion magnitude, and
  observation geometry. We also establish the observability condition,
  approximation validity boundaries, Cram\'er--Rao lower bound, and a
  provably consistent bias-eliminated estimator
  (Sections~\ref{sec:be}--\ref{sec:crlb}).
  \item Extensive synthetic and real-world experiments establish new
  state-of-the-art accuracy while achieving the fastest runtime among all
  evaluated methods. With five points and 0.5~pixel noise, the monocular
  and binocular solvers achieve translation RMSEs of 3.70~mm and 3.91~mm,
  rotation RMSEs of $10.09''$ (combined pitch--yaw) and $10.58''$, and
  runtimes of 0.34~ms and 0.27~ms, respectively.
\end{enumerate}

\section{Related Work}
\subsection{Perspective-n-Point (PnP)}
The Perspective-$n$-Point (PnP) problem, which determines camera pose from
$N$ 3D-2D point correspondences, has been extensively studied over the past
decades. It is broadly categorized into the minimal three-point case (P3P)
and the overdetermined case ($N>3$). The P3P problem requires a minimum of
three point correspondences to determine the camera pose, typically
yielding up to four geometrically feasible solutions. Classical approaches
introduce the point-to-camera-center distances as intermediate unknowns,
leading to a system of quadratic equations. A complete analytical solution
was provided by Gao \emph{et al.}~\cite{r16} using the Wu--Ritt zero
decomposition. Direct methods, which estimate the pose without an
intermediate alignment step, have gained popularity in recent years owing
to their simpler pipeline and better numerical stability. Kneip \emph{et
al.}~\cite{r14} proposed a direct geometric P3P solver with improved
efficiency. Ke and Roumeliotis~\cite{r17} derived a compact algebraic
formulation free of numerically risky computations such as tangent
operations, achieving state-of-the-art stability. Nakano~\cite{r18}
presented a simple direct solution by exploiting the algebraic structure of
the problem. More recently, degenerate-conic-based methods have emerged as
an alternative to classical quartic-based formulations. Lambda
Twist~\cite{r19} demonstrated that the P3P problem can be reduced to a
univariate cubic equation via the intersection of two conics, offering
benefits in efficiency and avoidance of duplicate solutions. Ding \emph{et
al.}~\cite{r20} revisited the P3P problem and reinforced the
degenerate-conic framework with a more compact analysis of the conic-cubic
relationship. Zhang \emph{et al.}~\cite{r21} proposed a direct
degenerate-conic-based solver that finds the degeneration through in-plane
rotations, providing geometric interpretations to the degenerate conics.

For the overdetermined PnP problem ($N>3$), methods are generally formulated as least-squares optimization problems. Early iterative approaches, such as the method of Lu \emph{et al.}~\cite{r22}, refined pose estimates through alternating minimization but were susceptible to local minima. A significant breakthrough came with EPnP~\cite{r13}, an $O(n)$ non-iterative method that represents 3D points using four virtual control points and solves for their coordinates via null space analysis. This linear formulation inspired subsequent improvements. Li \emph{et
al.}~\cite{r23} proposed RPnP, a robust $O(n)$ solution based on the
construction of a rational function. Wang
\emph{et al.}~\cite{r48} subsequently proposed SRPnP, a simple and
efficient solver that robustly handles ordinary 3D, planar, and
quasi-singular point configurations. Hesch and Roumeliotis~\cite{r24}
formulated the Direct Least-Squares (DLS) method that minimizes a nonlinear
object-space error through global polynomial solving. Zheng \emph{et
al.}~\cite{r25} revisited the PnP problem and derived a fast, general, and
optimal solution by solving the first-order optimality conditions via the
Gr\"obner basis technique. The pursuit of globally optimal solutions
continued with several notable contributions. Zhou and Kaess~\cite{r26}
presented an efficient and accurate algorithm (EOPnP) that exploits null
space analysis of the linear system and enforces the orthonormality
constraints of the rotation matrix to achieve global optimality. Terzakis
and Lourakis~\cite{r27} proposed SQPnP, a consistently fast and globally
optimal method that applies sequential quadratic programming from multiple
initializations. Lourakis and Terzakis~\cite{r28} further achieved global
optimality using the modified Rodrigues parameterization (MRP).

Iterative methods generally provide higher accuracy at the expense of
additional computation. Classical Gauss--Newton refinement or motion-only
bundle adjustment (BA)~\cite{r4} minimizes the reprojection error on the
manifold of $SO(3)$, but requires a reliable initialization to avoid
convergence to local minima. Ferraz \emph{et al.}~\cite{r29} proposed
REPPnP, which integrates algebraic outlier rejection within an efficient
PnP framework. Garro \emph{et al.}~\cite{r30} formulated PPnP as an
anisotropic orthogonal Procrustes problem solved via block relaxation. More
recently, Zhou \emph{et al.}~\cite{r31} proposed an iterative PnP solver
that approximates the reprojection cost using second-order polynomials and
determines the optimal step size analytically, improving accuracy for large
depth ranges. The combination of a non-iterative initialization followed by
iterative refinement remains a widely adopted paradigm for balancing
accuracy and computational cost. Several recent works have pushed the
boundaries of PnP methods in various directions. Zeng \emph{et
al.}~\cite{r32} proposed CPnP, which subtracts the asymptotic bias of a
closed-form solution to yield a provably consistent estimate. Sun \emph{et
al.}~\cite{r33} introduced ACEPnP, integrating geometry constraints into
the control-point formulation via quadratically constrained quadratic
programming. Wu \emph{et al.}~\cite{r34} provided a globally optimal
framework for quadratic pose estimation problems with solvability analysis.
Zhan \emph{et al.}~\cite{r15} proposed GMLPnP, a maximum likelihood solver
considering anisotropic observation uncertainty via iterated generalized
least squares, decoupled from the camera model. Henry and
Christian~\cite{r35} developed optimal DLT-based PnP solutions. Zeng
\emph{et al.}~\cite{r36} extended the bias-elimination idea to stereo
visual odometry for consistent large-scale localization.

\subsection{Generalized Perspective-n-Point (GPnP)}
The generalized PnP (gPnP) problem extends classical PnP to multi-camera
systems, where 3D-2D correspondences may originate from different cameras
with known relative poses. It is similarly divided into the minimal
three-point case (gP3P) and overdetermined methods.

The gP3P problem admits up to eight solutions. Early methods employed
geometric formulations to derive an eighth-order univariate
polynomial~\cite{r37,r38}. Later works shifted toward algebraic approaches,
typically arriving at a system of three quadratic equations in three
unknowns (3Q3), which is then reduced to a univariate polynomial via the Sylvester
resultant~\cite{r39}, polynomial eigenvalue~\cite{r40}, or specialized
elimination~\cite{r41}. The Gr\"obner basis method has also been applied
for polynomial solving~\cite{r42}.

For the overdetermined gPnP configuration, Schweighofer and
Pinz~\cite{r43} proposed an early SDP-based iterative method, though it
suffered from low accuracy and convergence to local optima. Kneip \emph{et
al.}~\cite{r42} introduced GPnP, adopting the core ideas of EPnP,
specifically, null space analysis on virtual control points, thereby
inheriting EPnP's efficiency but also its sensitivity to point distribution
and noise. Moreover, GPnP degenerates under planar configurations. Kneip
\emph{et al.}~\cite{r44} proposed UPnP, which directly formulates
constraints based on the non-perspective model without relying on
intermediate variables, and solves the first-order optimality conditions
via the Gr\"obner basis, achieving higher accuracy. Wientapper \emph{et
al.}~\cite{r45} improved upon UPnP with GAPS, employing an automatically
generated Gr\"obner basis solver and a modified formulation, though both
UPnP and GAPS remain computationally demanding due to the complexity of
their resulting polynomial systems. Zhang \emph{et al.}~\cite{r46} proposed
EA-GPnP, a null-space-based gPnP method that achieves high accuracy and efficiency through an optimized formulation and two specialized polynomial solvers. 
Campos \emph{et al.}~\cite{r47} presented POSEAMM, a unified framework that handles diverse pose problems through alternating minimization.

\begin{figure*}[t]
\centering
\includegraphics[width=\textwidth]{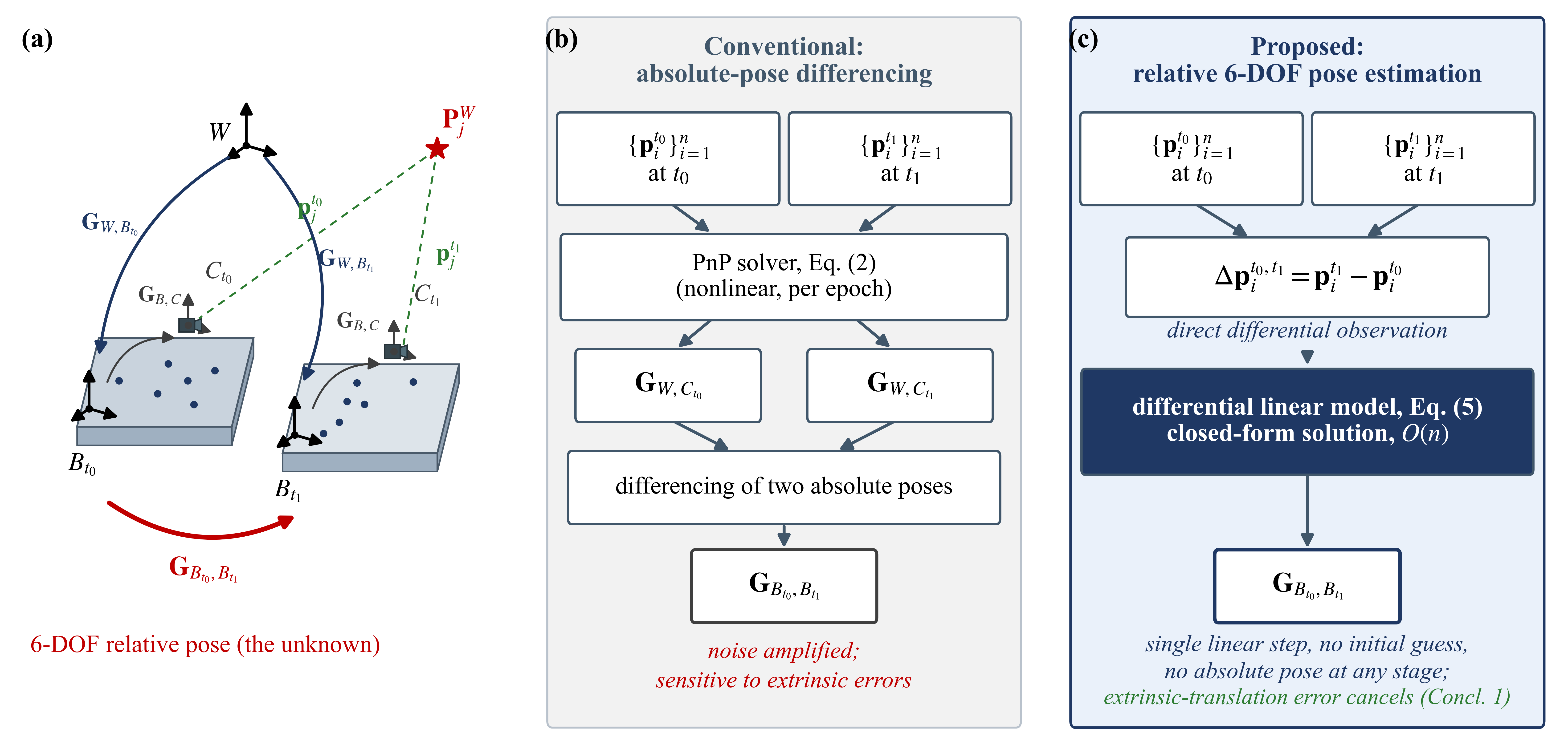}
\caption{Schematic diagram of the PnP-like method and the proposed method for the relative pose estimation: (a) scene and coordinate frames; (b) the conventional pipeline solves two absolute poses and differences them; (c) the proposed pipeline takes the image-point differences as the direct observation and solves the relative pose in a single linear step, with no absolute pose solved at any stage.}
\label{fig:schematic}
\end{figure*}

It should be noted that the pose estimated by the aforementioned PnP and
gPnP methods represents the rigid transformation from the world frame to
the camera frame. To obtain the relative pose of the measurement platform,
one typically needs to solve the camera pose at two different time instants
independently and then recover the platform motion through coordinate
transformation, as illustrated in Fig.~\ref{fig:schematic}. This two-step
pipeline applies to both monocular and multi-camera configurations. In
contrast, the method proposed in this work bypasses the per-frame pose
estimation step entirely and directly constructs a differential model for
relative pose estimation from 2D point differences and 3D coordinates,
which will be described in detail in the next section.

\emph{Learning-based pipelines.} Learned correspondence front ends such as SuperPoint~\cite{superpoint} and SuperGlue~\cite{superglue} are complementary to the proposed method because they can provide the 2D observations consumed by the estimator. The formulation itself is agnostic to how correspondences are obtained. End-to-end pose regressors in the PoseNet family~\cite{posenet} are less directly compatible with the differential formulation because their prediction errors need not be temporally correlated. Consequently, differencing two predictions does not guarantee that their biases will cancel. We therefore retain a geometric estimation layer.

\section{Principle and Method}
\subsection{Transformation Between Coordinate Systems}
The relationship between two coordinate systems can be represented by a rigid
transformation matrix. As shown in Fig.~\ref{fig:schematic}(a), we consider
three coordinate systems: the world frame ($W$), the camera frame ($C$), and
the platform frame ($B$). The platform and camera frames at times $t_0$ and
$t_1$ are denoted by $B_{t_0}$, $B_{t_1}$, $C_{t_0}$, and $C_{t_1}$,
respectively. The relationships among these frames are
\begin{equation}
\label{eq:frames}
\begin{aligned}
\bm{P}^{B} &= \bm{G}_{W,B}\,\bm{P}^{W}, &
\bm{P}^{C} &= \bm{G}_{B,C}\,\bm{P}^{B},\\
\bm{P}^{B_{t_1}} &= \bm{G}_{B_{t_0},B_{t_1}}\bm{P}^{B_{t_0}}, &
\bm{P}^{C_{t_1}} &= \bm{G}_{C_{t_0},C_{t_1}}\bm{P}^{C_{t_0}},
\end{aligned}
\end{equation}
where the 3D point $\bm{P}=[\bm{\bar{P}}^{T}\;1]^{T}$ is in the form of
homogeneous coordinates with $\bm{\bar{P}}=[X\;Y\;Z]^{T}$,
$\bm{G}_{A,B}=\bigl[\begin{smallmatrix}\bm{R}_{A,B}&\bm{T}_{A,B}\\
\bm{0}_{1\times 3}&1\end{smallmatrix}\bigr]$ denotes the rigid
transformation matrix, $\bm{R}_{A,B}$ and $\bm{T}_{A,B}$ are the rotation
matrix and translation vector from coordinate system $A$ to $B$, and
$\bm{G}_{B_{t_0},B_{t_1}}$ denotes the relative 6-DOF pose of the
measurement platform, which contains the parameters that need to be
estimated in this paper. In addition, the camera is fixed on the platform,
and then $\bm{G}_{B_{t_0},C_{t_0}}=\bm{G}_{B_{t_1},C_{t_1}}=\bm{G}_{B,C}$.

\subsection{Perspective Projection}
\begin{figure}[htbp]
  \centering
  \includegraphics[
    width=\columnwidth,
    trim=18bp 45bp 55bp 10bp,
    clip
  ]{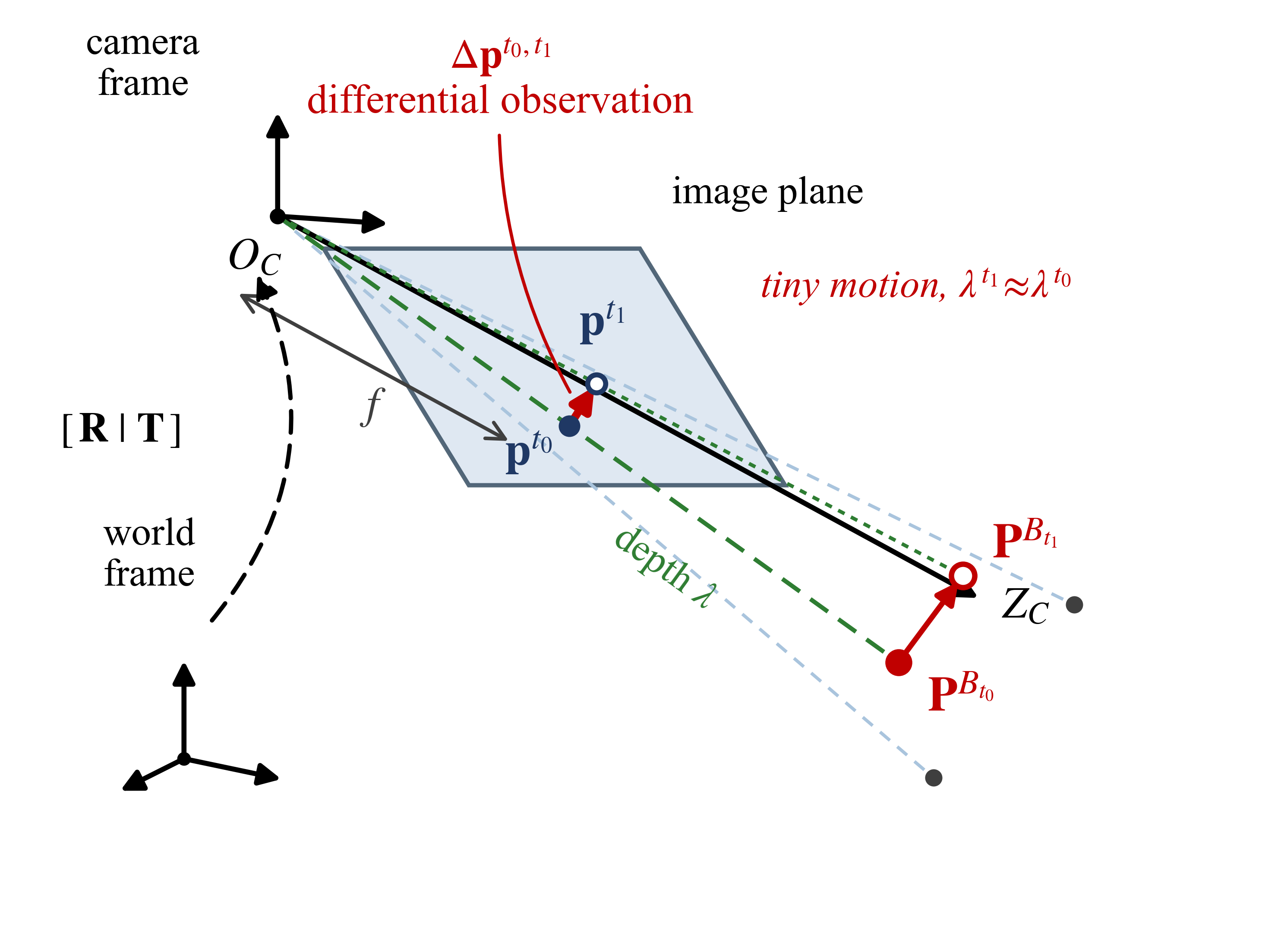}
  \caption{Perspective projection model under small platform motion. The
  observed point moves to $\bm{P}^{B_{t_1}}$ while its depth remains
  approximately unchanged, $\lambda^{t_1}\approx\lambda^{t_0}$. The image-plane
  displacement $\Delta\bm{p}\dt$ forms the differential observation.}
  \label{fig:projection}
  \end{figure}

Throughout this paper, a calibrated camera with a perspective projection
model is utilized. As shown in Fig.~\ref{fig:projection}, the projection of a 3D world point $\bm{P}^{W}$ onto the image plane is
\begin{equation}
\label{eq:projection}
\lambda\,\bm{p}=\left[\bm{K}\;\;\bm{0}_{3\times 1}\right]
\bm{G}_{W,C}\,\bm{P}^{W},
\end{equation}
where the 2D image point $\bm{p}$ and 3D world point $\bm{P}^{W}$ are in
the form of homogeneous coordinates,
$\bm{K}=\bigl[\begin{smallmatrix}f_x&0&c_x\\0&f_y&c_y\\0&0&1
\end{smallmatrix}\bigr]$, and $\lambda$ denotes the depth factor of
$\bm{P}^{W}$. In the ideal case, the projection of the 3D point must
coincide with the image point. When noise is present in the measurement
data, we denote $\tilde{\bm{p}}$ as the noisy observation of the projection
of the 3D point $\bm{P}^{W}$.

\subsection{Proposed Algorithm}
\label{sec:algorithm}
According to \eqref{eq:projection}, the relationship between the 3D world
point $\bm{P}^{W}$ and the image point $\bm{p}$ at time $t_0$ and $t_1$ can
be expressed as
\begin{equation}
\label{eq:twoepoch}
\begin{aligned}
\lambda^{t_0}\bm{p}^{t_0}
 &=\left[\bm{K}\;\;\bm{0}_{3\times 1}\right]
   \bm{G}_{B_{t_0},C_{t_0}}\bm{G}_{W,B_{t_0}}\bm{P}^{W}\\
 &=\left[\bm{K}\;\;\bm{0}_{3\times 1}\right]
   \bm{G}_{B,C}\,\bm{G}_{W,B_{t_0}}\bm{P}^{W},\\
\lambda^{t_1}\bm{p}^{t_1}
 &=\left[\bm{K}\;\;\bm{0}_{3\times 1}\right]
   \bm{G}_{B,C}\,\bm{G}_{B_{t_0},B_{t_1}}\bm{G}_{W,B_{t_0}}\bm{P}^{W}.
\end{aligned}
\end{equation}
In \eqref{eq:twoepoch}, the intrinsic parameter matrix $\bm{K}$ and the
extrinsic parameter matrix $\bm{G}_{B,C}$ of the camera can be calibrated
before the pose estimation procedure. If the measurement platform at time
$t_0$ is taken as the reference, the 3D world point $\bm{P}^{W}$ can be
expressed in $B_{t_0}$ as $\bm{P}^{B_{t_0}}$. Then, \eqref{eq:twoepoch} can
be rewritten as
\begin{equation}
\label{eq:refframe}
\begin{aligned}
\lambda^{t_0}\bm{p}^{t_0}
 &=\left[\bm{K}\;\;\bm{0}_{3\times 1}\right]\bm{G}_{B,C}\,
   \bm{P}^{B_{t_0}},\\
\lambda^{t_1}\bm{p}^{t_1}
 &=\left[\bm{K}\;\;\bm{0}_{3\times 1}\right]\bm{G}_{B,C}\,
   \bm{G}_{B_{t_0},B_{t_1}}\bm{P}^{B_{t_0}}.
\end{aligned}
\end{equation}
In practice, especially for outdoor measurement scenarios, the value of
$\lambda^{t_1}$ is generally tens or even hundreds of meters, whereas the
change of $\lambda$ induced by the measurement platform motion is
negligible. Hence, the value of $\lambda$ is assumed to be constant in this
paper; in other words, $\lambda^{t_1}$ for the 3D point $\bm{P}^{B_{t_0}}$
at time $t_1$ can be set to $\lambda^{t_0}$. Subtracting the two equations
in \eqref{eq:refframe}, we have
\begin{equation}
\label{eq:diffmodel}
\begin{aligned}
\lambda^{t_0}\bigl(\bm{p}^{t_1}-\bm{p}^{t_0}\bigr)
 &=\left[\bm{K}\;\;\bm{0}_{3\times 1}\right]\bm{G}_{B,C}
   \bigl(\bm{G}_{B_{t_0},B_{t_1}}-\bm{E}\bigr)\bm{P}^{B_{t_0}},\\
\Delta\bm{p}\dt
 &=\frac{1}{\lambda^{t_0}}
   \left[\bm{K}\;\;\bm{0}_{3\times 1}\right]\bm{G}_{B,C}
   \bigl(\bm{G}_{B_{t_0},B_{t_1}}-\bm{E}\bigr)\bm{P}^{B_{t_0}},
\end{aligned}
\end{equation}
where $\bm{E}$ is a $4\times 4$ identity matrix.

Assuming that the motion of the platform is tiny, the Lie group formulation
is adopted to represent the rigid transformation from the platform frame at
time $t_0$ to the platform frame at time $t_1$, and then
$\bm{G}_{B_{t_0},B_{t_1}}$ can be expressed as
\begin{equation}
\label{eq:expmap}
\bm{G}_{B_{t_0},B_{t_1}}=\exp(\bm{\eta})
 =\exp\Bigl(\sum_{j=0}^{5}\eta_j\bm{G}_j\Bigr)
 \approx\bm{E}+\sum_{j=0}^{5}\eta_j\bm{G}_j,
\end{equation}
where $\bm{\eta}=[\bm{\eta}_T^{T}\;\;\bm{\eta}_R^{T}]^{T}$, $\bm{\eta}_T$
and $\bm{\eta}_R$ are the motion velocities corresponding to translations
in the $X$, $Y$, $Z$ directions and rotations about the $X$, $Y$, $Z$ axes
respectively, and $\bm{G}_j$ are the group generators, as follows:
\newcommand{\genm}[1]{\Bigl[\begin{smallmatrix}#1\end{smallmatrix}\Bigr]}%
\begin{equation}
\label{eq:generators}
\begin{aligned}
\bm{G}_0&=\genm{0&0&0&1\\0&0&0&0\\0&0&0&0\\0&0&0&0},&
\bm{G}_1&=\genm{0&0&0&0\\0&0&0&1\\0&0&0&0\\0&0&0&0},&
\bm{G}_2&=\genm{0&0&0&0\\0&0&0&0\\0&0&0&1\\0&0&0&0},\\
\bm{G}_3&=\genm{0&0&0&0\\0&0&-1&0\\0&1&0&0\\0&0&0&0},&
\bm{G}_4&=\genm{0&0&1&0\\0&0&0&0\\-1&0&0&0\\0&0&0&0},&
\bm{G}_5&=\genm{0&-1&0&0\\1&0&0&0\\0&0&0&0\\0&0&0&0}.
\end{aligned}
\end{equation}
Substituting \eqref{eq:expmap} into \eqref{eq:diffmodel}, we have
\begin{equation}
\label{eq:genform}
\Delta\bm{p}\dt=\frac{1}{\lambda^{t_0}}
 \left[\bm{K}\;\;\bm{0}_{3\times 1}\right]\bm{G}_{B,C}
 \sum_{j=0}^{5}\eta_j\bm{G}_j\,\bm{P}^{B_{t_0}}.
\end{equation}
If we use the rotation matrix and translation vector to express the transformation from the platform frame at time $t_0$ to the platform frame at time $t_1$, \eqref{eq:genform} can be rewritten as
\begin{equation}
\label{eq:skewform}
\begin{aligned}
\Delta\bm{p}\dt
 &=\frac{1}{\lambda^{t_0}}\bm{K}\bm{R}_{B,C}
   \bigl(\skewm{\bm{\eta}_R}\bm{P}^{B_{t_0}}+\bm{\eta}_T\bigr)\\
 &=\frac{1}{\lambda^{t_0}}\bm{K}\bm{R}_{B,C}
   \bigl(-\skewm{\bm{P}^{B_{t_0}}}\bm{\eta}_R+\bm{\eta}_T\bigr)\\
 &=\frac{1}{\lambda^{t_0}}\bm{K}\bm{R}_{B,C}
   \bigl[\bm{I}_{3\times 3}\;\;-\skewm{\bm{P}^{B_{t_0}}}\bigr]\bm{\eta},
\end{aligned}
\end{equation}
where $\skewm{\bm{\eta}_R}$ is the corresponding skew-symmetric matrix of vector $\bm{\eta}_R$, $\bm{M}=\frac{1}{\lambda^{t_0}}\bm{K}\bm{R}_{B,C} \bigl[\bm{I}_{3\times 3}\;\;-\skewm{\bm{P}^{B_{t_0}}}\bigr]$, and $\Delta\bm{p}\dt$ denotes the motion of the projection of the world point $\bm{P}$ on the image plane from time $t_0$ to time $t_1$. Given $n$ 3D-2D point correspondences with $n\ge 3$, we have
\begin{equation}
\label{eq:normal}
\bm{M}\bm{\eta}=\bm{P},
\end{equation}
where $\bm{M}=[\bm{M}_1^{T}\;\cdots\;\bm{M}_n^{T}]^{T}$ and
$\bm{P}=[(\Delta\bm{p}_1\dt)^{T}\;\cdots\;(\Delta\bm{p}_n\dt)^{T}]^{T}$.

Equation~\eqref{eq:normal} can be solved by the least squares method. Moreover, according to \eqref{eq:genform}, the objective function with respect to the
motion parameters $\bm{\eta}$ of the measurement platform can be expressed as follows:
\begin{equation}
\label{eq:objective}
\begin{aligned}
E(\bm{\eta})=\min\sum_{i=1}^{n}\Bigl\|
 \Delta\bm{p}_i\dt-\frac{1}{\lambda_i^{t_0}}&
 \left[\bm{K}\;\,\bm{0}_{3\times 1}\right]\bm{G}_{B,C}\\[-2pt]
 &\cdot\sum_{j=0}^{5}\eta_j\bm{G}_j\,\bm{P}_i^{B_{t_0}}\Bigr\|_2^{2}.
\end{aligned}
\end{equation}
The relative 6-DOF pose can be solved by the LM method according to
\eqref{eq:objective}. The partial derivative of $\Delta\bm{p}\dt$ with
respect to the $j$-th generating motion $\eta_j$ can be computed as
\begin{equation}
\label{eq:derivative}
\frac{\partial\Delta\bm{p}_i\dt}{\partial\eta_j}
 =\frac{1}{\lambda_i^{t_0}}
  \left[\bm{K}\;\;\bm{0}_{3\times 1}\right]\bm{G}_{B,C}\,
  \bm{G}_j\,\bm{P}_i^{B_{t_0}}.
\end{equation}

When the intrinsic parameter matrix of the camera is known, the poses
$\bm{G}_{B_{t_0},C_{t_0}}$ and $\bm{G}_{B_{t_0},C_{t_1}}$ from the platform
frame at time $t_0$ to the camera frames at times $t_0$ and $t_1$ can be
obtained by solving \eqref{eq:refframe}, according to PnP-like methods.
Next, the pose change of the platform coordinate system between the two
consecutive times, i.e., the 6-DOF relative pose of the platform, can be
obtained. In addition, the pose $\bm{G}_{B,C}$ at the initial time can be
solved by the PnP-like methods first, and then
$\bm{G}_{B_{t_0},B_{t_1}}$ can be directly solved by the second equation
of \eqref{eq:refframe}.

The main difference between the proposed method and the PnP-like methods is
that the depth factor invariance hypothesis (i.e., the invariance of
$\lambda$) is introduced in the proposed method. The benefits will be
analyzed in detail in the next section. A quantitative validity boundary of
this hypothesis --- an explicit bound on the measurable axial displacement
--- is established in Section~\ref{sec:crlb}
(Proposition~\ref{prop:bounds} and
Eqs.~\eqref{eq:depthresid}--\eqref{eq:depthdecomp}).

\subsection{Theoretical Analysis}
\label{sec:theory}
In this section, the influence of extrinsic parameter calibration error on
the relative pose estimation is analyzed. If the calibration errors for the
extrinsic parameters are small, $\bm{G}_{\mathrm{error}}$ can also be
represented in the Lie group formulation as
\begin{equation}
\label{eq:gerror}
\bm{G}_{\mathrm{error}}=\exp(\bm{\mu})
 =\exp\Bigl(\sum_{j=0}^{5}\mu_j\bm{G}_j\Bigr)
 \approx\bm{E}+\sum_{j=0}^{5}\mu_j\bm{G}_j,
\end{equation}
where $\bm{\mu}=[\bm{\mu}_T^{T}\;\;\bm{\mu}_R^{T}]^{T}$. Substituting
\eqref{eq:gerror} into \eqref{eq:diffmodel}, the equation can be rewritten
as
\begin{equation}
\label{eq:perturbed}
\begin{aligned}
\Delta\bm{p}\dt
 &=\frac{1}{\lambda^{t_0}}\left[\bm{K}\;\;\bm{0}_{3\times 1}\right]
   \bm{G}_{\mathrm{error}}\bm{G}_{B,C}
   \bigl(\bm{G}_{B_{t_0},B_{t_1}}-\bm{E}\bigr)\bm{P}^{B_{t_0}}\\
 &=\frac{1}{\lambda^{t_0}}\left[\bm{K}\;\;\bm{0}_{3\times 1}\right]
   \Bigl(\bm{E}+\sum_{j=0}^{5}\mu_j\bm{G}_j\Bigr)\\
 &\qquad\cdot\bm{G}_{B,C}
   \sum_{j=0}^{5}\eta_j\bm{G}_j\,\bm{P}^{B_{t_0}}\\
 &=\frac{1}{\lambda^{t_0}}\left[\bm{K}\;\;\bm{0}_{3\times 1}\right]
   \bigl(\Delta\bm{G}_1+\Delta\bm{G}_2\bigr)\bm{P}^{B_{t_0}},
\end{aligned}
\end{equation}
where $\Delta\bm{G}_1=\bm{G}_{B,C}\sum_{j=0}^{5}\eta_j\bm{G}_j$ is
independent of the extrinsic parameter calibration error, and
$\Delta\bm{G}_2=\sum_{j=0}^{5}\mu_j\bm{G}_j\,\bm{G}_{B,C}
\sum_{j=0}^{5}\eta_j\bm{G}_j$ involves both the extrinsic parameter
calibration error and the relative pose of the measurement platform. By
comparing the matrix norms of $\Delta\bm{G}_1$ with those of
$\Delta\bm{G}_2$, we can draw the following conclusions.

\begin{conclusionx}[Exact immunity to translational extrinsic calibration
errors]
\label{con:translation}
$\Delta\bm{G}_2$ is immune to the translational extrinsic calibration
errors in $\bm{G}_{\mathrm{error}}$ ($\mu_0,\mu_1,\mu_2$), and related only
to the rotation errors in $\bm{G}_{\mathrm{error}}$ ($\mu_3,\mu_4,\mu_5$).
Consequently, within the proposed model, errors in the calibrated
camera-to-platform translation do not perturb the estimated platform
motion; the remaining calibration-dependent perturbation contains only the
rotational component of the extrinsic error. See Appendix~A of the
supplemental material for a complete proof.
\end{conclusionx}

\begin{conclusionx}[Bounded sensitivity to rotational extrinsic calibration
errors]
\label{con:range}
The rotational extrinsic calibration error is closely related to the
measurement range of the platform's relative pose. The rotational extrinsic
calibration error, the measurement resolution, and the maximum measurement
range all have an impact on the measurement results. By setting an
appropriate measurement range, the impact of the camera's calibration error
can be ignored. Similarly, if the camera's calibration error tends to zero,
the range approaches infinity, which corresponds to the PnP-like methods.
See Appendix~B of the supplemental material for a complete proof.
\end{conclusionx}

The above analysis addresses the monocular configuration. It can be
extended to the multi-camera configuration along the same line. Suppose the
measurement platform carries $m$ rigidly mounted cameras, and the extrinsic
parameters of the $k$-th camera are calibrated with a tiny error, which is
represented in the Lie group formulation as
\begin{equation}
\label{eq:multierror}
\bm{G}_{\mathrm{error}}^{k}
 =\exp\Bigl(\sum_{j=0}^{5}\mu_j^{k}\bm{G}_j\Bigr)
 \approx\bm{E}+\sum_{j=0}^{5}\mu_j^{k}\bm{G}_j,\qquad k=1,\dots,m,
\end{equation}
where the error twist of the $k$-th camera is defined as in
\eqref{eq:gerror}. Each camera contributes a stacked measurement block of
the form of \eqref{eq:normal}, and the multi-camera relative pose is given
by the joint normal equations
\begin{equation}
\label{eq:jointnormal}
\bm{\eta}=\Bigl(\sum_{k=1}^{m}\tilde{\bm{M}}_k^{T}\tilde{\bm{M}}_k
 \Bigr)^{-1}\sum_{k=1}^{m}\tilde{\bm{M}}_k^{T}\bm{P}_k,\qquad
\tilde{\bm{M}}_k=\bm{M}_k+\Delta\bm{M}_k.
\end{equation}
Substituting \eqref{eq:multierror} into each block and retaining the
first-order terms, the perturbed coefficient matrix of each camera splits
into a calibration-independent part and a perturbation term possessing
exactly the same structure as that of the monocular case. Accordingly, the
first-order estimation error of the relative pose is
\begin{equation}
\label{eq:firstorder}
\delta\bm{\eta}\approx
 \Bigl(\sum_{k=1}^{m}\bm{M}_k^{T}\bm{M}_k\Bigr)^{-1}
 \sum_{k=1}^{m}\bm{M}_k^{T}\Delta\bm{M}_k\,\bm{\eta}.
\end{equation}

\begin{conclusionx}[Translation immunity of the multi-camera model]
\label{con:multitrans}
The perturbation term of each camera is independent of the translational
extrinsic calibration errors of all the cameras and involves only its own
rotation calibration error; the proof of Conclusion~\ref{con:translation} in
Appendix~A of the supplemental material applies to each camera block
separately, and no cross terms between different cameras arise. Hence the
multi-camera differential model
completely inherits the immunity to the translation calibration errors.
\end{conclusionx}

\begin{conclusionx}[Attenuation of rotational-error effects by
complementary views]
\label{con:multirot}
The multi-camera configuration further attenuates the influence of the
rotational extrinsic calibration errors through two mechanisms. First, according to
\eqref{eq:firstorder} the perturbations of the $m$ cameras are combined by
a weighted average; when the rotation calibration errors of different
cameras are independent and zero-mean, the expected pose error decreases at
the rate of the square root of the number of cameras. Second, the amplification
of the perturbation is bounded by the inverse of the smallest singular
value of the stacked coefficient matrix; complementary viewing directions
(e.g., two cameras whose optical axes form a large included angle) raise
this singular value substantially --- in particular for the rotation
component about the optical axis, which is nearly unobservable for a single
camera with a narrow field of view --- so that the same calibration
perturbation maps to a much smaller pose error. Moreover, the bilinear
structure between the rotation calibration error and the platform motion is
preserved, and thus for micro-motion measurement the calibration-induced
error of the multi-camera solution remains a second-order small quantity.
These conclusions are consistent with the simulation results in
Section~\ref{sec:simbino}.
\end{conclusionx}

\subsection{Bias-Eliminated Consistent Estimation}
\label{sec:be}
The least-squares (LS) solution of \eqref{eq:normal} is statistically reliable only when the stacked coefficient matrix is noise-free.
In practice, the matrix is constructed from measured quantities, and the resulting errors-in-variables problem~\cite{tls,fuller} makes the LS estimator inconsistent because its bias does not vanish as the number of  points increases.
This section characterizes the bias in closed form and derives a bias-eliminated (BE) estimator that
restores consistency. Throughout, centered image coordinates
$\tilde{u}=u-c_x$, $\tilde{v}=v-c_y$ are used, and per point
$\bm{M}_i=\bm{\Pi}_i\bm{A}_i$ with
$\bm{\Pi}_i=\frac{1}{\lambda_i}
\bigl[\begin{smallmatrix}f&0&-\tilde{u}_i\\0&f&-\tilde{v}_i
\end{smallmatrix}\bigr]$ and
$\bm{A}_i=\bigl[\bm{I}_3\;\;-\skewm{\bm{P}_i^{B}}\bigr]$.

The third column of $\bm{\Pi}_i$ is the dominant error source. The pixel offset $(\tilde{u}_i,\tilde{v}_i)$ is obtained from the image measurement at $t_0$, whose noise $\bm{e}_{0,i}$ is correlated with the differential  observation $\Delta\bm{p}_i=\bm{p}_{1,i}-\bm{p}_{0,i}$.
Writing the constructed matrix as $\tilde{\bm{M}}_i=\bm{M}_i+\Delta\bm{M}_i$, a direct
expansion gives
\begin{equation}
\label{eq:dM}
\Delta\bm{M}_i=-\frac{1}{\lambda_i}\bm{e}_{0,i}\bm{a}_i^{T},\qquad
\bm{a}_i=[0\;\;0\;\;1\;\;Y_i\;\;-X_i\;\;0]^{T},
\end{equation}
where $\bm{e}_{0,i}$ is zero-mean with covariance $\sigma^{2}\bm{I}_2$ and
$(X_i,Y_i)$ are the lateral coordinates of $\bm{P}_i^{B}$. Since the
observation noise is $\bm{\varepsilon}_i=\bm{e}_{1,i}-\bm{e}_{0,i}$, the
second-order moments follow by direct computation:
\begin{equation}
\label{eq:moments}
\mathbb{E}\bigl[\tilde{\bm{M}}_i^{T}\tilde{\bm{M}}_i\bigr]
 =\bm{M}_i^{T}\bm{M}_i+\frac{2\sigma^{2}}{\lambda_i^{2}}
  \bm{a}_i\bm{a}_i^{T},\qquad
\mathbb{E}\bigl[\tilde{\bm{M}}_i^{T}\bm{\varepsilon}_i\bigr]
 =\frac{2\sigma^{2}}{\lambda_i}\bm{a}_i.
\end{equation}
The first term of \eqref{eq:moments} inflates the normal matrix (an
attenuation-type effect), while the second introduces a nonzero correlation
between the regressor and the noise --- the source of the asymptotic bias~\cite{fuller}.
Notably, $\bm{a}_i$ spans exactly the depth-perturbing directions --- its
inner product with the motion vector equals the depth rate of point $i$ ---
so after inversion by the ill-conditioned normal matrix the LS bias
concentrates on the weakly observable axial translation, as verified in
Section~\ref{sec:bevalid}. Both correction terms are computable from known
quantities up to the noise level $\sigma^{2}$; subtracting them from the
empirical moments yields the bias-eliminated estimator
\begin{equation}
\label{eq:be}
\bm{\eta}_{\mathrm{BE}}
 =\Bigl(\sum_i\Bigl[\tilde{\bm{M}}_i^{T}\tilde{\bm{M}}_i
  -\frac{2\sigma^{2}}{\lambda_i^{2}}\bm{a}_i\bm{a}_i^{T}\Bigr]\Bigr)^{-1}
  \sum_i\Bigl[\tilde{\bm{M}}_i^{T}\Delta\bm{p}_i
  -\frac{2\sigma^{2}}{\lambda_i}\bm{a}_i\Bigr],
\end{equation}
where $\sigma^{2}$ is any consistent estimate of the noise level~\cite{r32,r36}, e.g., the
residual-based
\begin{equation}
\label{eq:sigmahat}
\hat{\sigma}^{2}
 =\frac{\bigl\|\bm{P}-\tilde{\bm{M}}
   \bm{\eta}_{\mathrm{LS}}\bigr\|_2^{2}}{2(2n-6)}
\end{equation}
(each coordinate of $\bm{\varepsilon}_i$ has variance $2\sigma^{2}$). The
corrected solve costs the same as ordinary LS.

\begin{theorem}[Consistency and asymptotic normality]
\label{thm:consistency}
Assume the noise terms are i.i.d., zero-mean with finite fourth moments,
the points and depths are bounded, and
$\frac{1}{n}\sum_i\bm{M}_i^{T}\bm{M}_i\to\bm{Q}\succ 0$. Then (i) the LS
estimate converges to a biased limit,
$\bm{\eta}_{\mathrm{LS}}\to\bm{\eta}+\bm{\beta}$ with
$\bm{\beta}=2\sigma^{2}(\bm{Q}+2\sigma^{2}\bm{D})^{-1}
(\bm{d}-\bm{D}\bm{\eta})$, where
$\bm{D}=\lim\frac{1}{n}\sum\bm{a}_i\bm{a}_i^{T}/\lambda_i^{2}$ and
$\bm{d}=\lim\frac{1}{n}\sum\bm{a}_i/\lambda_i$; (ii) the BE estimate is
consistent and asymptotically normal:
$\sqrt{n}(\bm{\eta}_{\mathrm{BE}}-\bm{\eta})$ converges in distribution to
a zero-mean Gaussian~\cite{vaart}. See Appendix~C of the supplemental
material for the proof.
\end{theorem}

\begin{remark}[Surveyed-coordinate errors]
If the 3D coordinates carry survey errors with known covariance, an
analogous correction applies, with $\Delta\bm{M}_i$ given by the Jacobian
of $\bm{M}_i$ with respect to $\bm{P}_i^{B}$; the correction terms remain
closed-form and the two corrections can be combined.
\end{remark}

\begin{remark}[Asymptotic efficiency]
\label{rem:efficiency}
A single Gauss--Newton iteration of \eqref{eq:objective} initialized at
$\bm{\eta}_{\mathrm{BE}}$ is asymptotically efficient, i.e., it attains the
Cram\'er--Rao lower bound of Section~\ref{sec:crlb}, by the classical
one-step-estimator argument~\cite{vaart}.
\end{remark}

\subsection{Observability, the Cram\'er--Rao Lower Bound, and Modeling
Boundaries}
\label{sec:crlb}
We now characterize when the differential system \eqref{eq:normal} is
solvable and how accurately any estimator can solve it.

\begin{theorem}[Observability]
\label{thm:observability}
Let $\bm{M}$ be the $2n\times 6$ stacked matrix of \eqref{eq:normal}.
(i) $n\ge 3$ is necessary for $\operatorname{rank}(\bm{M})=6$. (ii) If all
points lie on a line with direction $\bm{d}$ through a point $\bm{q}$, then
$\operatorname{rank}(\bm{M})\le 5$: the screw motion
$\bm{\eta}_0=[-\bm{d}\times\bm{q};\;\bm{d}]$ gives every point the velocity
$\bm{d}\times(\bm{P}_i-\bm{q})=\bm{0}$ and thus annihilates every
observation row. (iii) If the $n\ge 3$ points are not collinear,
$\operatorname{rank}(\bm{M})=6$ for all configurations outside a
measure-zero algebraic set; for $n=3$ this exceptional set reduces to the
classical singularity cylinder through the three points known from the P3P
literature. See Appendix~D of the supplemental material.
\end{theorem}

\begin{theorem}[CRLB and anisotropy]
\label{thm:crlb}
For Gaussian noise $\bm{\varepsilon}_i$ with covariance
$2\sigma^{2}\bm{I}_2$, any unbiased estimator of $\bm{\eta}$ satisfies~\cite{kay}
\begin{equation}
\label{eq:crlb}
\operatorname{Cov}(\hat{\bm{\eta}})\succeq\bm{F}^{-1}
 =2\sigma^{2}\Bigl(\sum_i\bm{M}_i^{T}\bm{M}_i\Bigr)^{-1}.
\end{equation}
Moreover, if all image points lie within radius $\rho_{\max}$ of the
principal point, the ratio between the bound's weak-mode (roll, axial
translation) and strong-mode (pitch/yaw, lateral translation) standard
deviations is at least $f/\rho_{\max}=\cot(\mathrm{FOV}/2)$ up to
point-distribution constants. Combined with
Remark~\ref{rem:efficiency}, the BE estimator followed by one
Gauss--Newton step attains the bound asymptotically. See Appendix~D of the
supplemental material.
\end{theorem}

\begin{proposition}[Modeling-error boundaries]
\label{prop:bounds}
The two modeling approximations of Section~\ref{sec:algorithm} obey exact
bounds. (i) Depth invariance: substituting the exact identity
$\Delta\tilde{u}=\frac{f}{\lambda_0}\Delta X
-\frac{\Delta\lambda}{\lambda_0}\tilde{u}_1$ shows the induced pixel
residual equals $\frac{\Delta\lambda}{\lambda}(\bm{p}_1-\bm{c})$
identically; it is buried in noise iff
$|T_Z|\le\lambda\sigma/\rho_{\max}$. (ii) First-order truncation: the
single-linearization residual is $\frac{1}{2}f\theta^{2}$ (bound
$\theta\le\sqrt{2\sigma/f}$), reduced to at most $O(\theta^{3})$
(coefficient $\frac{1}{6}f$) by one re-linearization --- consistent with
the $\pm 30'$ working range reported in Section~\ref{sec:experiments}.
\end{proposition}

The depth-invariance bound in Proposition~\ref{prop:bounds}(i) rests on an
exact identity rather than a Taylor expansion, and it can be turned into an
explicit working range for the platform motion. Adding and subtracting
$fX_1/\lambda_0$ in the projection difference of one image coordinate
yields, without any approximation, the residual discarded by the
depth-invariance hypothesis in vector form:
\begin{equation}
\label{eq:depthresid}
\bm{\varepsilon}_d=\Delta\bm{p}_{\mathrm{model}}
 -\Delta\bm{p}_{\mathrm{true}}
 =\frac{\Delta\lambda}{\lambda_0}\bigl(\bm{p}_1-\bm{c}\bigr),\qquad
\|\bm{\varepsilon}_d\|=\frac{|\Delta\lambda|}{\lambda}\,\rho,
\end{equation}
where $\bm{c}$ is the principal point and $\rho=\|\bm{p}_1-\bm{c}\|$ the
pixel eccentricity. The discarded term is thus a pure radial ``zoom'' of
the image about the principal point: a point at the principal point
($\rho=0$) is unaffected, and the field corner ($\rho=\rho_{\max}$) is the
worst case, which is why the bound is stated at $\rho_{\max}$. The depth
change over the differencing interval decomposes along the optical axis as
\begin{equation}
\label{eq:depthdecomp}
\Delta\lambda=\bm{e}_3^{T}\bm{R}_{B,C}
 \bigl(\bm{t}-\skewm{\bm{P}^{B}}\bm{\theta}\bigr)
 =T_Z+O\Bigl(\frac{\rho}{f}\lambda\|\bm{\theta}\|\Bigr),
\end{equation}
where $\bm{e}_3$ is the optical-axis unit vector and $(\bm{t},\bm{\theta})$
are the translational and rotational components of $\bm{\eta}$. The
rotation-induced term in \eqref{eq:depthdecomp} carries the small factor
$\rho/f$ and stays below 1~mm over the $\pm 30'$ working range at
$\lambda=100$~m, so the axial platform translation $T_Z$ dominates the
depth change. Requiring the worst-case residual of \eqref{eq:depthresid} to
be buried in the image noise, $(|\Delta\lambda|/\lambda)\rho_{\max}\le
\sigma$, recovers the measurable-axial-displacement range
$|T_Z|\le\lambda\sigma/\rho_{\max}$ of Proposition~\ref{prop:bounds}(i).
Under the long-range monitoring configuration of
Section~\ref{sec:experiments} ($\lambda=100$~m, $\sigma=0.2$~pixel,
$\rho_{\max}\approx 2200$ pixels for a $3840\times 2160$ sensor) the range
evaluates to $|T_Z|\le 9$~mm at the field corner and relaxes to about
20~mm at mid-field ($\rho\approx 1000$ pixels). Millimeter-level axial
platform motion therefore leaves no systematic bias visible above the noise
--- consistent with the flat axial-translation error curves observed in the
simulations --- while the linear growth of the residual with $|T_Z|$ beyond
the bound marks the transition to the iterative exact-projection variant.

\section{Experimental Results}
\label{sec:experiments}
In this section, the proposed methods are validated by comparison with
several state-of-the-art algorithms, including LHM~\cite{r22},
EPnP+GN~\cite{r13}, RPnP~\cite{r23}, DLS~\cite{r24}, ASPnP~\cite{r49}, and
OPnP~\cite{r25}. All algorithms were executed and tested in MATLAB.

\subsection{Numerical Simulations for Monocular Vision}
\label{sec:simmono}
The simulated configuration reproduces the long-range bridge-monitoring
scenario: the camera resolution is $3840\times 2160$ pixels with an
equivalent focal length of 100{,}000 pixels; the control points are
randomly distributed at 50--100~m from the camera without perturbation; the
translational extrinsic calibration error is within $\pm 5$~mm; the three
attitude angles of the measurement station vary within $\pm 30$~arcmin and
the translation components within $\pm 1$~mm, and the full 6-DOF relative
pose is estimated. The default simulation parameters are listed in
Table~\ref{tab:params}. The proposed monocular differential linear solution
(with the two-pass re-linearization) is compared with the representative
PnP methods LHM, EPnP+GN, RPnP, DLS, ASPnP, and OPnP, which estimate the
absolute poses at the two instants and difference them to obtain the
relative pose; each condition is repeated 100 times (200 for the
typical-condition statistics), and trials whose rotation error exceeds
$1^{\circ}$ are counted as divergent and excluded. All reported runtimes
are measured in MATLAB R2024b on a 13th-Gen Intel Core i7-1360P processor.

\begin{table}[htbp]
\centering
\caption{Default Simulation Parameters}
\label{tab:params}
\begin{tabular}{ll}
\toprule
Parameter & Value\\
\midrule
Camera resolution & $3840\times 2160$ pixels\\
Equivalent focal length & 100{,}000 pixels\\
Control point distance & 50--100 m\\
Number of control points & 4--50 (default 5)\\
Image localization noise & 0--1 pixel (default 0.2, typical 0.5)\\
Attitude change & within $\pm 30$ arcmin\\
Translation & within $\pm 1$ mm\\
Extrinsic translation calib. error & $\pm 5$ mm\\
Monte-Carlo trials & 100 (200 for typical condition)\\
\bottomrule
\end{tabular}
\end{table}

\subsubsection{Influence of the Number of Control Points}
\begin{figure*}[!tb]
\centering
\includegraphics[width=\textwidth]{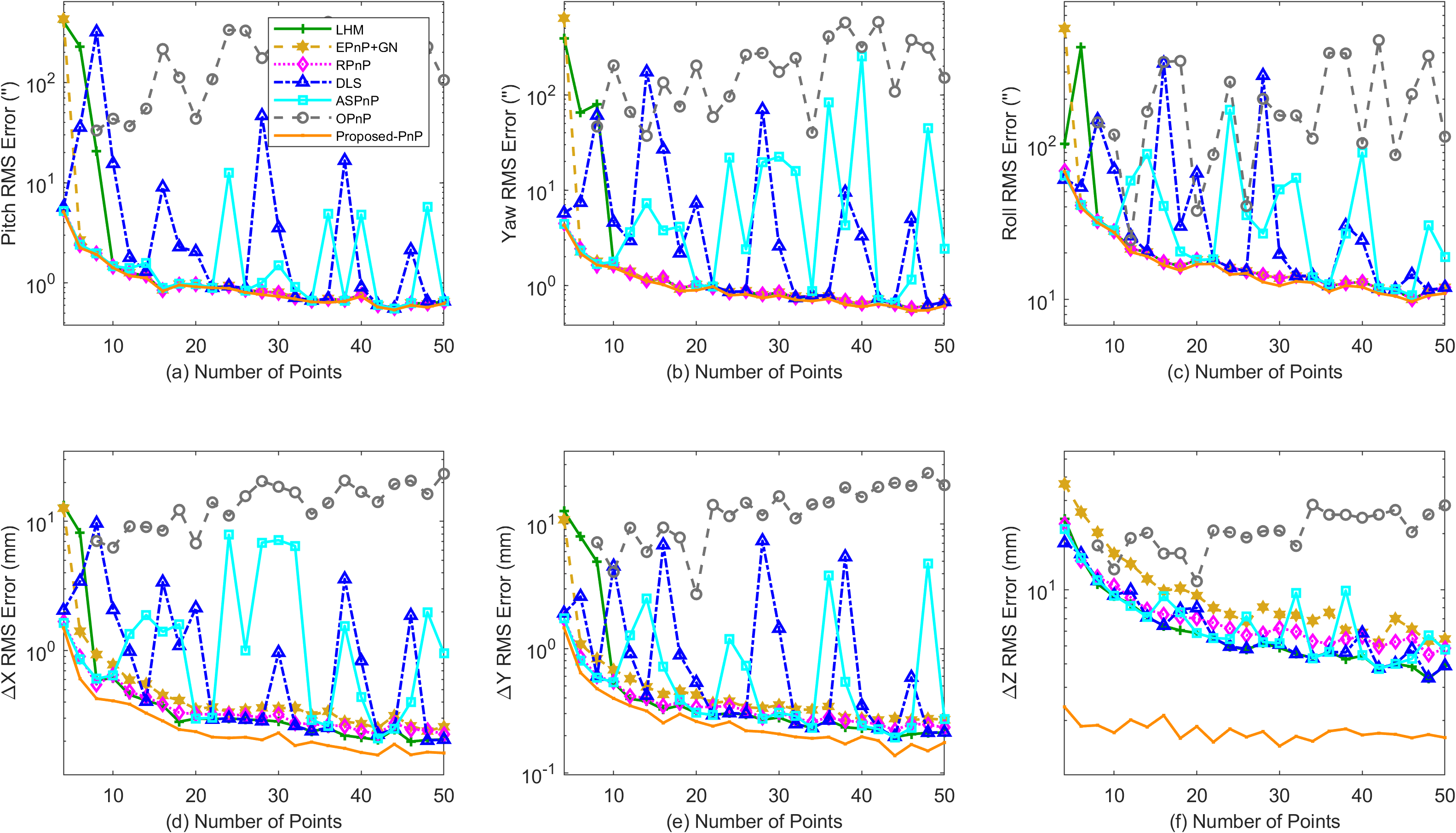}
\caption{RMSE of the rotation and translation components w.r.t.\ the number
of control points: (a)--(c) rotation; (d)--(f) translation.}
\label{fig:mononpts}
\end{figure*}

Fig.~\ref{fig:mononpts} shows the RMSE of the three rotation and three translation components when the number of control points increases from 4 to 50 (0.2-pixel image noise). 
The proposed solution attains the lowest or joint-lowest error on every axis at every point count: with 4 points its pitch RMSE is $5.1''$ and its $\Delta X$ RMSE is 1.5~mm, whereas LHM reaches $404''$ and 13.3~mm, 
and OPnP fails completely with 4--6 points and diverges sporadically afterwards; with more points the proposed solution converges smoothly to $0.6''$ and 0.16~mm, without any divergence  throughout.

\subsubsection{Influence of Image Localization Error}
\begin{figure*}[!tb]
\centering
\includegraphics[width=\textwidth]{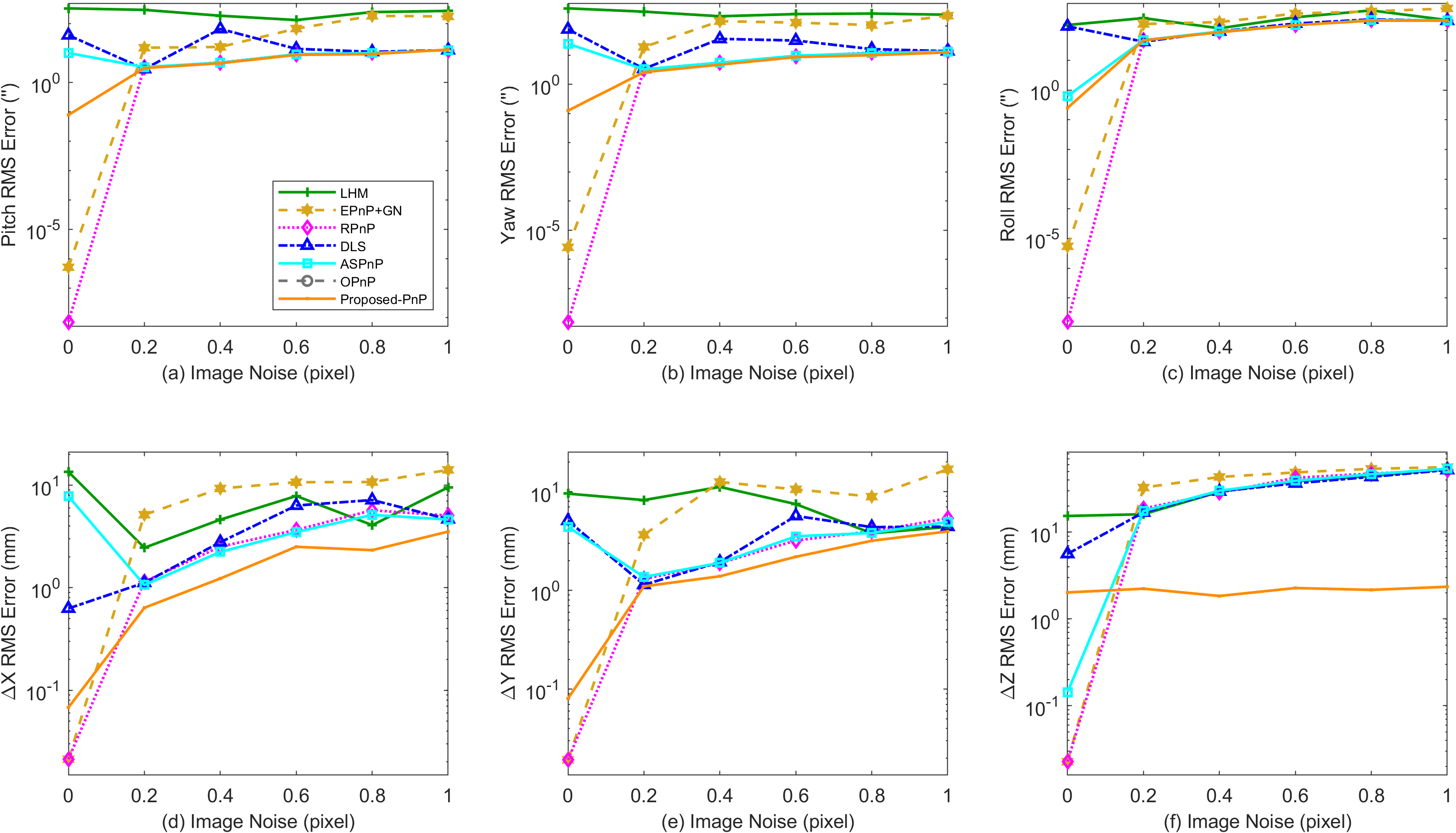}
\caption{RMSE of the rotation and translation components w.r.t.\ the image
localization noise: (a)--(c) rotation; (d)--(f) translation.}
\label{fig:mononoise}
\end{figure*}

Fig.~\ref{fig:mononoise} shows the results when the image localization
noise increases from 0 to 1 pixel (five control points). For the rotation,
the pitch/yaw RMSE of the proposed solution is comparable to or slightly
better than the best conventional method ($13.2''$/$12.4''$ at 1 pixel).
The advantage in translation is substantial, especially along the optical
axis: the $\Delta Z$ RMSE of the proposed solution stays at 2.0--2.4~mm
over the whole noise range, whereas that of the conventional methods grows
from about 19~mm to 52~mm --- more than one order of magnitude larger ---
because the depth direction is weakly observable in absolute pose
estimation and the differencing of two absolute solutions amplifies the
noise, which the differential formulation avoids by taking the inter-frame
image displacement as the direct observation. The proposed solution is also
the lowest in $\Delta X$/$\Delta Y$ (3.5~mm/4.0~mm at 1 pixel, versus
5.1~mm/5.4~mm for RPnP and 9.5~mm/4.4~mm for LHM).

\subsubsection{Influence of Extrinsic Calibration Errors}
\begin{figure*}[!tb]
\centering
\includegraphics[width=0.8\textwidth]{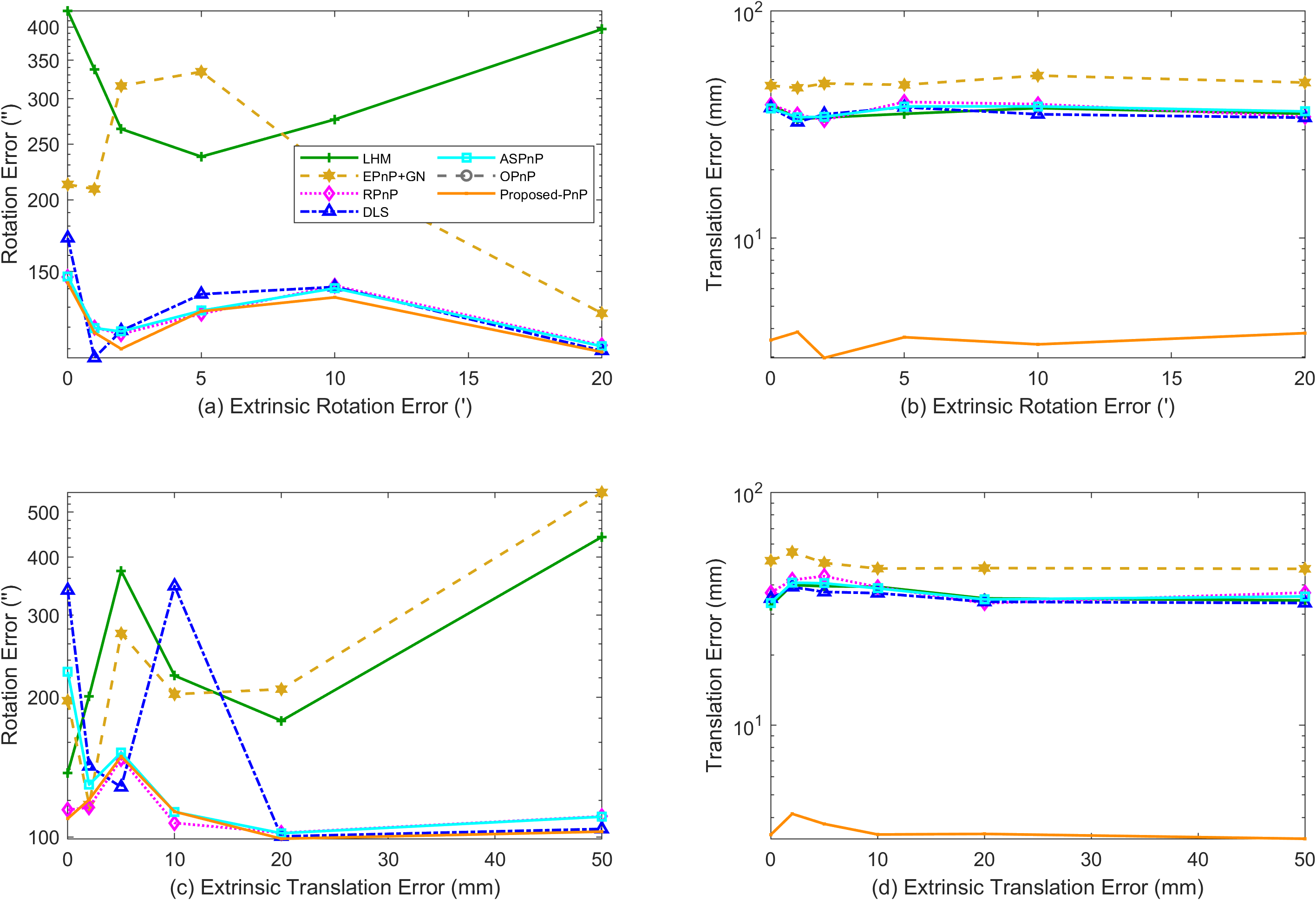}
\caption{Influence of the extrinsic calibration errors on the monocular
methods: (a)(b) extrinsic rotation error; (c)(d) extrinsic translation
error.}
\label{fig:monoextrinsic}
\end{figure*}

In practice the extrinsic parameters that relate the camera to the platform
are obtained by total-station-assisted calibration and inevitably contain
errors. Fig.~\ref{fig:monoextrinsic} examines the influence of the
extrinsic rotation error (0 to 20 arcmin, with the translation error fixed
at 5~mm) and of the extrinsic translation error (0 to 50~mm, with no
rotation error) on the monocular group. Varying the translational
extrinsic error produces no systematic degradation in the proposed
differential solution, empirically confirming its exact
translation-immunity property. By contrast, the rotational extrinsic error
introduces a nonzero but bounded perturbation whose magnitude remains
limited over the evaluated micro-motion range. Across the tested
calibration-error intervals, the translation RMSE of the proposed monocular
solution remains between 3.2 and 4.2~mm, while its rotation RMSE remains at
the noise-dominated level of approximately 110$''$--140$''$. The
comparatively flat curves of some frame-wise PnP baselines under this
particular configuration should not be interpreted as exact
calibration-error immunity, because their formulation does not eliminate
the translational extrinsic error algebraically.

\subsubsection{Statistical Distribution and Efficiency}
\begin{figure*}[!tb]
\centering
\includegraphics[width=0.8\textwidth]{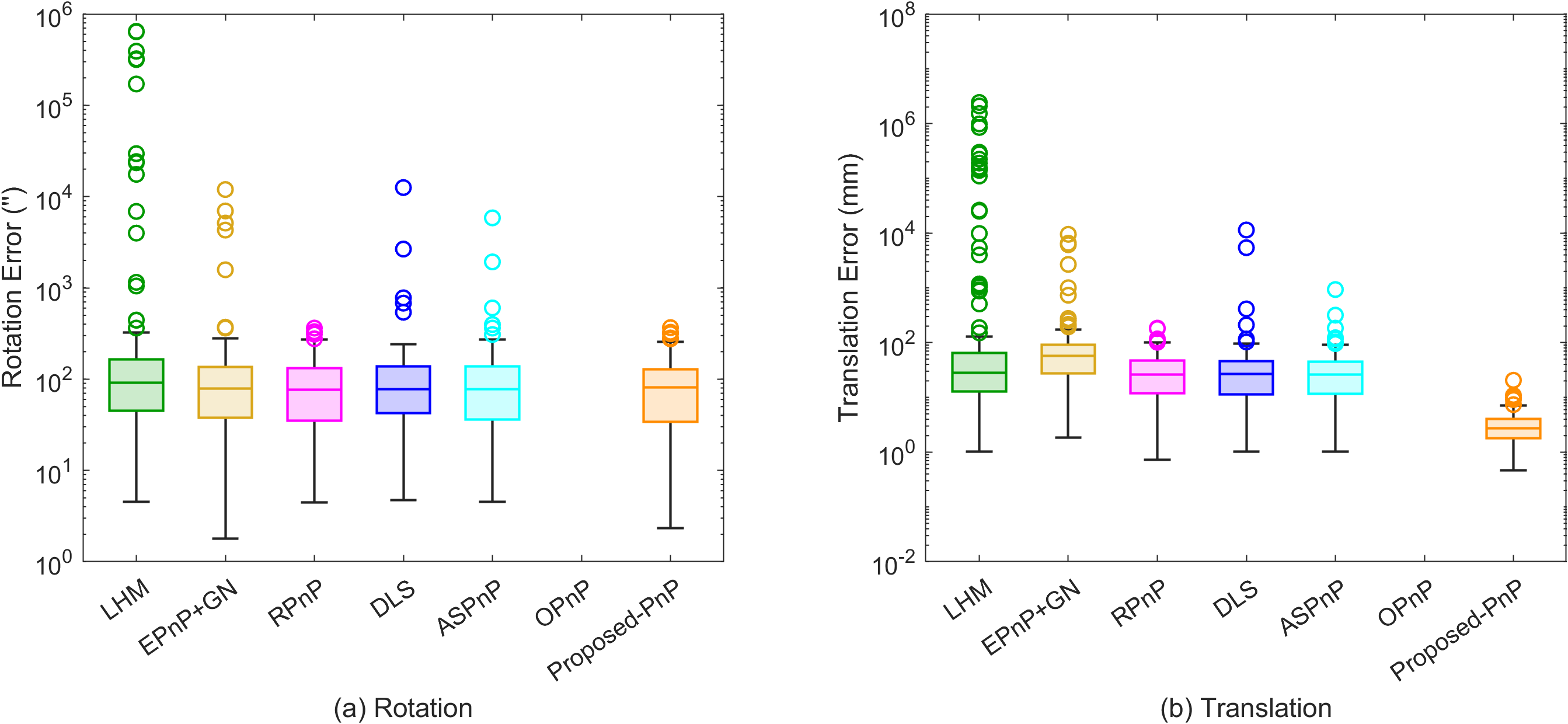}
\caption{Box plots of the rotation and translation errors under the typical
condition: (a) rotation; (b) translation.}
\label{fig:monobox}
\end{figure*}

\begin{figure}[htbp]
\centering
\includegraphics[width=0.9\columnwidth]{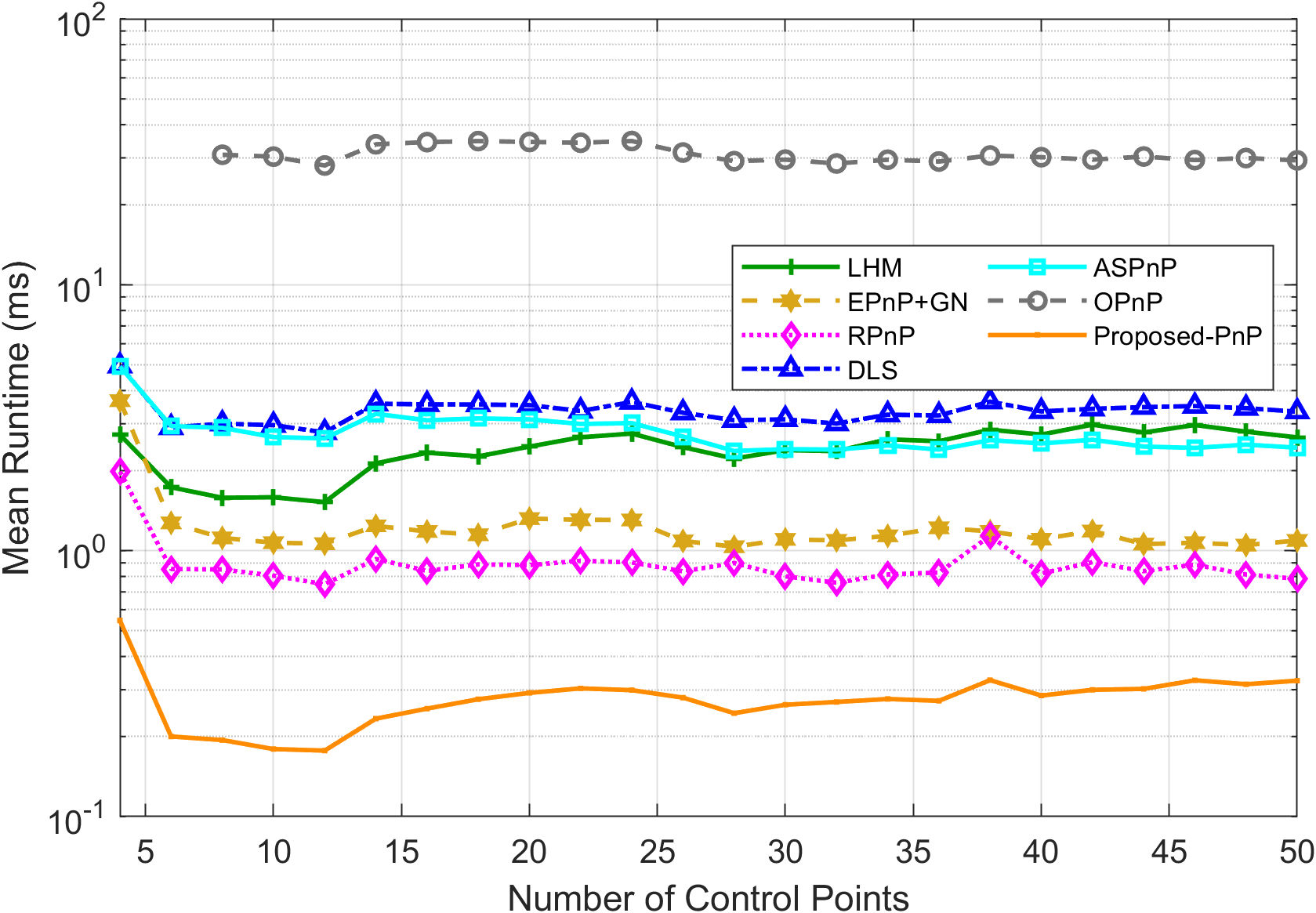}
\caption{Average runtime of the monocular methods versus the number of
control points (6-DOF estimation).}
\label{fig:monoruntime}
\end{figure}

Fig.~\ref{fig:monobox} depicts the error distributions under the typical condition (0.5-pixel noise, matching the reprojection error observed in the field experiment), and Fig.~\ref{fig:monoruntime} reports the average runtime. 
The rotation and translation boxes of the proposed solution are
the lowest in the group without heavy tails, with numerous outliers for LHM and OPnP. 
Statistically (200 trials, Table~\ref{tab:monostats}), the proposed solution attains the best translation RMSE of 3.70~mm (median 2.60~mm) and, in terms of the combined pitch--yaw error (the roll component is barely observable in the monocular geometry and is therefore excluded from the rotation statistics), the best rotation RMSE of $10.1''$ (median $6.5''$) with zero divergence. It takes only about 0.34~ms per solution, about $4\times$ faster than the fastest conventional method RPnP (1.4~ms) and more than an order of magnitude faster than OPnP.

\begin{table}[htbp]
\setlength{\tabcolsep}{1.0pt}
\centering
\caption{Rotation and Translation Error Statistics of the Monocular Methods
at the 0.5-Pixel Noise Level (5 Control Points; the Rotation Statistics
Refer to the Combined Pitch--Yaw Error, the Barely Observable Roll Being
Excluded)}
\label{tab:monostats}
\footnotesize
\setlength{\tabcolsep}{1.6pt}
\begin{tabular}{lccccc}
\toprule
Method & \makecell*[c]{Pitch--yaw\\RMSE/$''$}
       & \makecell*[c]{Pitch--yaw\\med./$''$}
       & \makecell*[c]{Trans.\\RMSE/mm} & \makecell*[c]{Trans.\\med./mm}
       & Time/ms\\
\midrule
LHM      & 407.20 & 7.73 & ---   & 32.70 & 2.20\\
EPnP+GN  & 29.50  & 7.65 & ---   & 44.56 & 2.52\\
RPnP     & 10.82  & 7.26 & 44.63 & 29.54 & 1.44\\
DLS      & 11.34  & 7.19 & ---   & 26.45 & 3.59\\
ASPnP    & 11.20  & 7.07 & 41.77 & 27.44 & 3.79\\
OPnP     & ---    & ---  & ---   & ---   & ---\\
Proposed-mono (linear) & \textbf{10.09} & \textbf{6.49} & \textbf{3.70}
 & \textbf{2.60} & \textbf{0.34}\\
\bottomrule
\end{tabular}
\end{table}

\subsection{Numerical Simulations for Binocular Vision}
\label{sec:simbino}
To further position the proposed method among the multi-camera
absolute-pose solvers reviewed in Section~II, the binocular differential
solutions (the proposed two-pass re-linearized linear solution and the
LM-refined solution, denoted BPnP) are compared with the generalized NPnP
solvers gOp~\cite{r43}, gDLS~\cite{r50}, UPnP~\cite{r44}, and
GAPS~\cite{r45}, as well as the recent efficient generalized solver
EA-GPnP~\cite{r46}. The simulated configuration follows the practical
constraints of the measurement platform: the equivalent focal length is
100{,}000 pixels with a $3840\times 2160$ resolution; five control points
are distributed at 50--100~m and observed by two cameras ($60^{\circ}$
between the optical axes, 0.6~m baseline); the three attitude angles vary
within $\pm 30$~arcmin, the translation components vary within $\pm 1$~mm,
and the extrinsic translation calibration error is within $\pm 5$~mm. The
control points are noise-free, the image localization noise increases from
0 to 1 pixel, and each condition is repeated 100 times. The generalized
solvers estimate the absolute rig pose at the two instants from the spatial
rays of both cameras and difference the two solutions to obtain the
relative pose of the platform.

\subsubsection{Influence of the Number of Control Points}
\begin{figure*}[!tb]
\centering
\includegraphics[width=\textwidth]{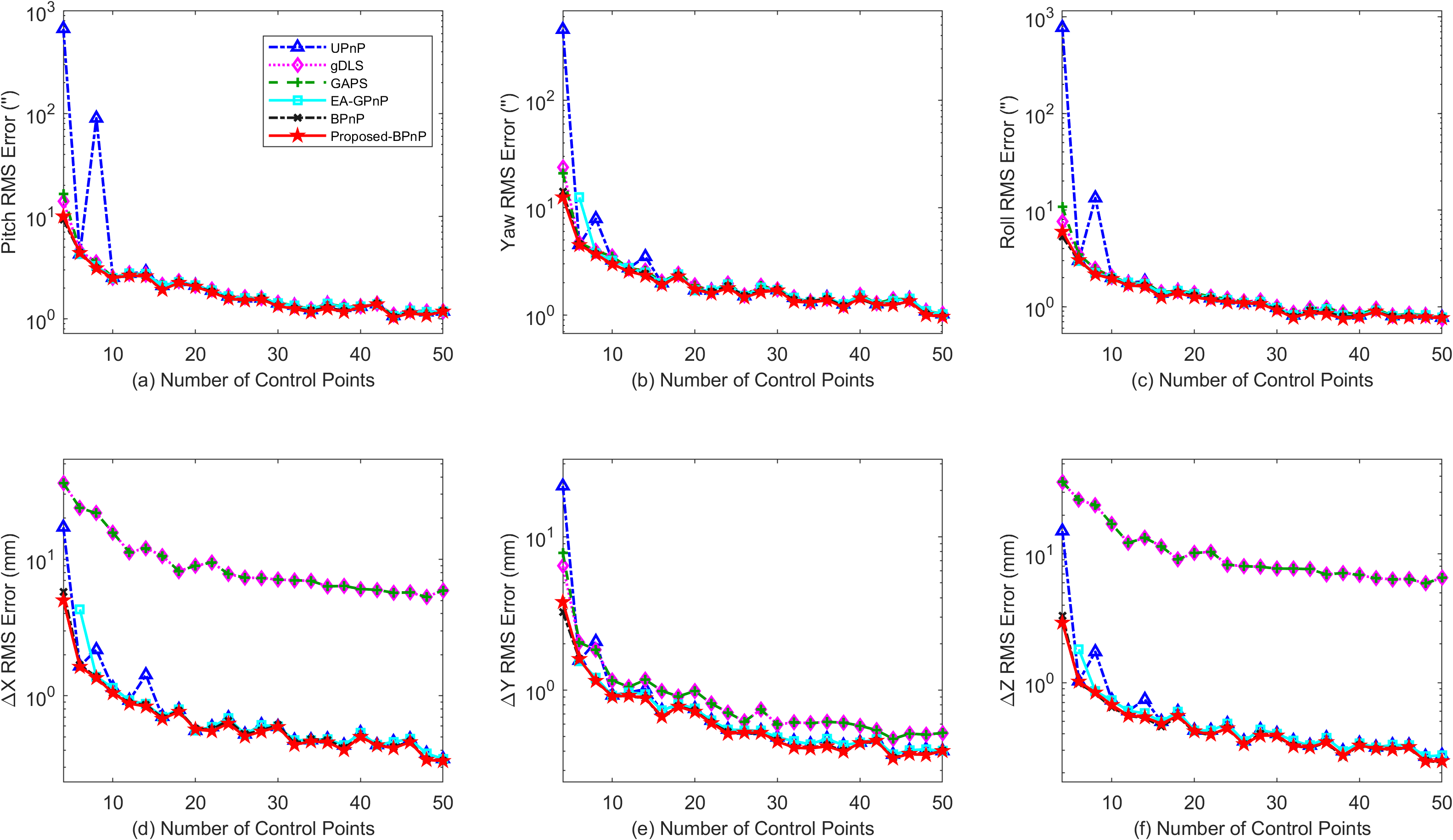}
\caption{RMSE of the rotation and translation components w.r.t.\ the number
of control points: (a)--(c) attitude angles; (d)--(f) translation
components.}
\label{fig:binonpts}
\end{figure*}

With the image noise fixed at 0.5 pixel, the number of control points
varies from 4 to 50 and each condition is repeated 100 times;
Fig.~\ref{fig:binonpts} shows the RMSE of the three attitude angles and of
the three translation components. The accuracy of the proposed linear
solution improves steadily with the number of points and coincides with the
LM-refined BPnP throughout: taking the depth direction as an example, the
$\Delta Z$ RMSE decreases from about 2.9~mm with 4 points to 0.25~mm with
50 points, whereas gDLS/GAPS converge from about 37~mm to only 6.5~mm and
remain an order of magnitude worse; UPnP suffers from sporadic divergence
at few points.

\subsubsection{Influence of Image Localization Error}
\begin{figure*}[!tb]
\centering
\includegraphics[width=\textwidth]{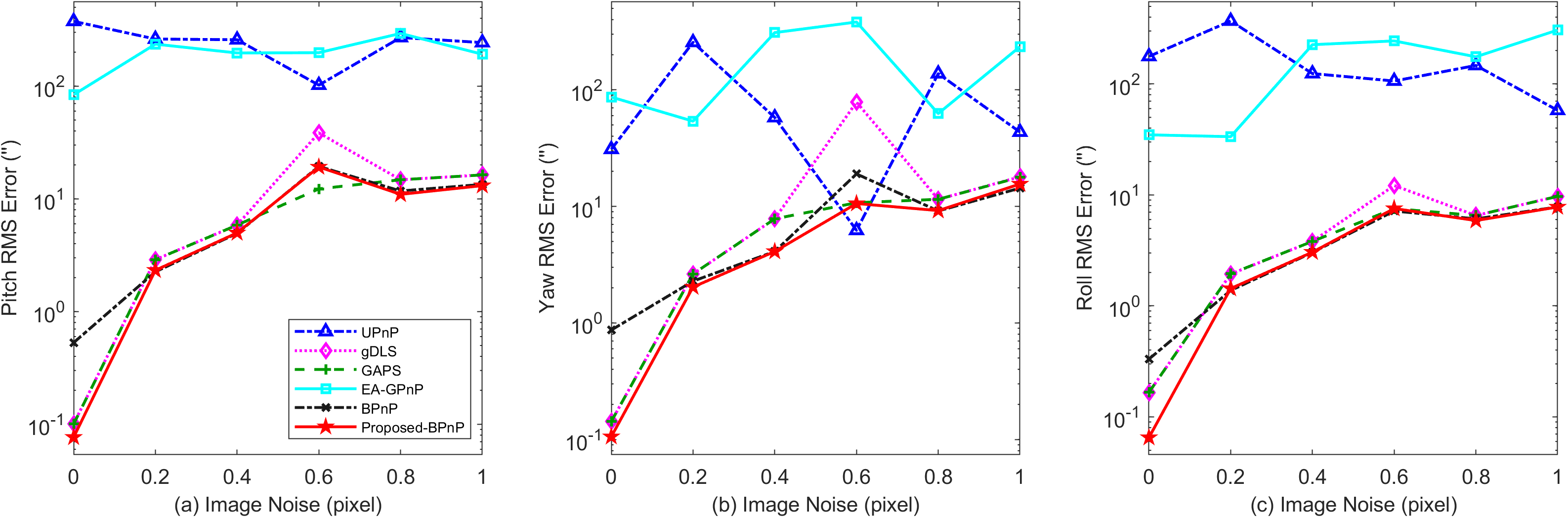}
\caption{RMSE of the three attitude angles w.r.t.\ the image localization
noise: (a) pitch; (b) yaw; (c) roll.}
\label{fig:binorot}
\end{figure*}

\begin{figure*}[!tb]
\centering
\includegraphics[width=\textwidth]{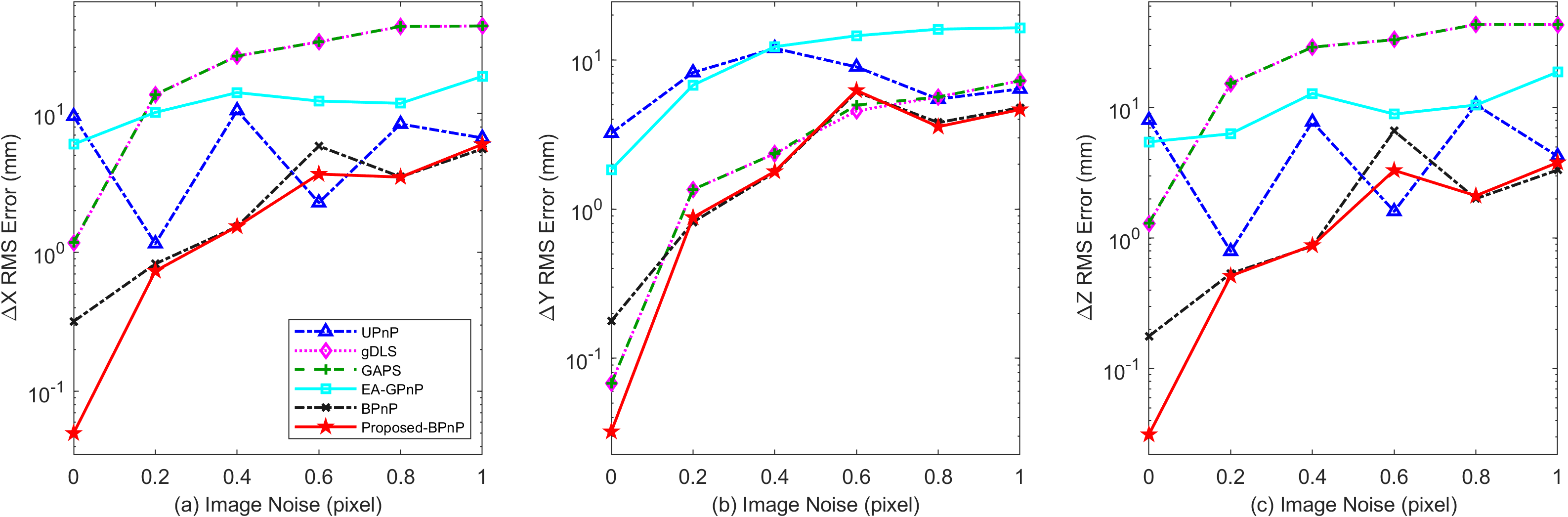}
\caption{RMSE of the three translation components w.r.t.\ the image
localization noise: (a) $X$; (b) $Y$; (c) $Z$.}
\label{fig:binotrans}
\end{figure*}

With five control points, the image localization noise increases from 0 to
1 pixel. Fig.~\ref{fig:binorot} and Fig.~\ref{fig:binotrans} show the RMSE
of the three attitude angles and of the three translation components,
respectively. The proposed linear solution and the LM-refined BPnP remain
the most accurate over the whole noise range, whereas the generalized
solvers exhibit substantially larger translation errors with heavier tails.

\subsubsection{Influence of Extrinsic Calibration Errors}
\begin{figure*}[!tb]
\centering
\includegraphics[width=0.8\textwidth]{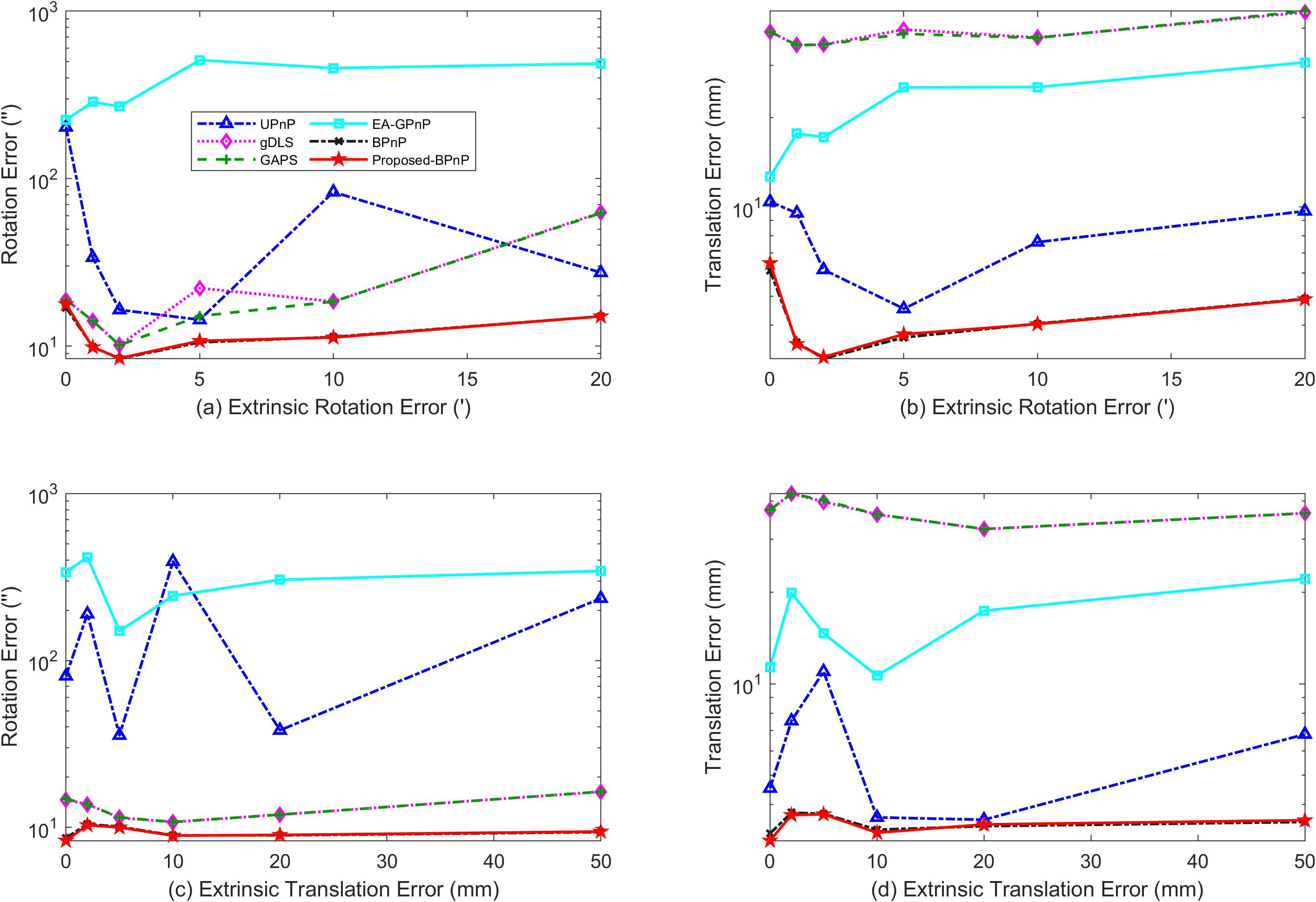}
\caption{Influence of the extrinsic calibration errors on the multi-camera
methods: (a)(b) extrinsic rotation error; (c)(d) extrinsic translation
error.}
\label{fig:binoextrinsic}
\end{figure*}

Fig.~\ref{fig:binoextrinsic} examines the influence of the extrinsic calibration errors on the multi-camera group, where the extrinsic rotation error increases from 0 to 20 arcmin (with the translation error fixed at 5~mm) and the extrinsic translation error increases from 0 to 50~mm (with no rotation error). 
The differential solutions are remarkably robust to both error sources. Translational extrinsic errors cancel exactly in the differential model, whereas rotational extrinsic errors enter through a bounded perturbation coupled to the platform motion. Consequently, the rotation and translation RMSEs of the proposed solution remain within $8$--$18''$ and 3.0--3.7~mm, respectively, over the evaluated range, with only mild degradation at the 20-arcmin extreme. This considerably relaxes the calibration accuracy required in field deployments.

\subsubsection{Influence of the Camera Mounting Angle}
\begin{figure*}[!tb]
\centering
\includegraphics[width=0.8\textwidth]{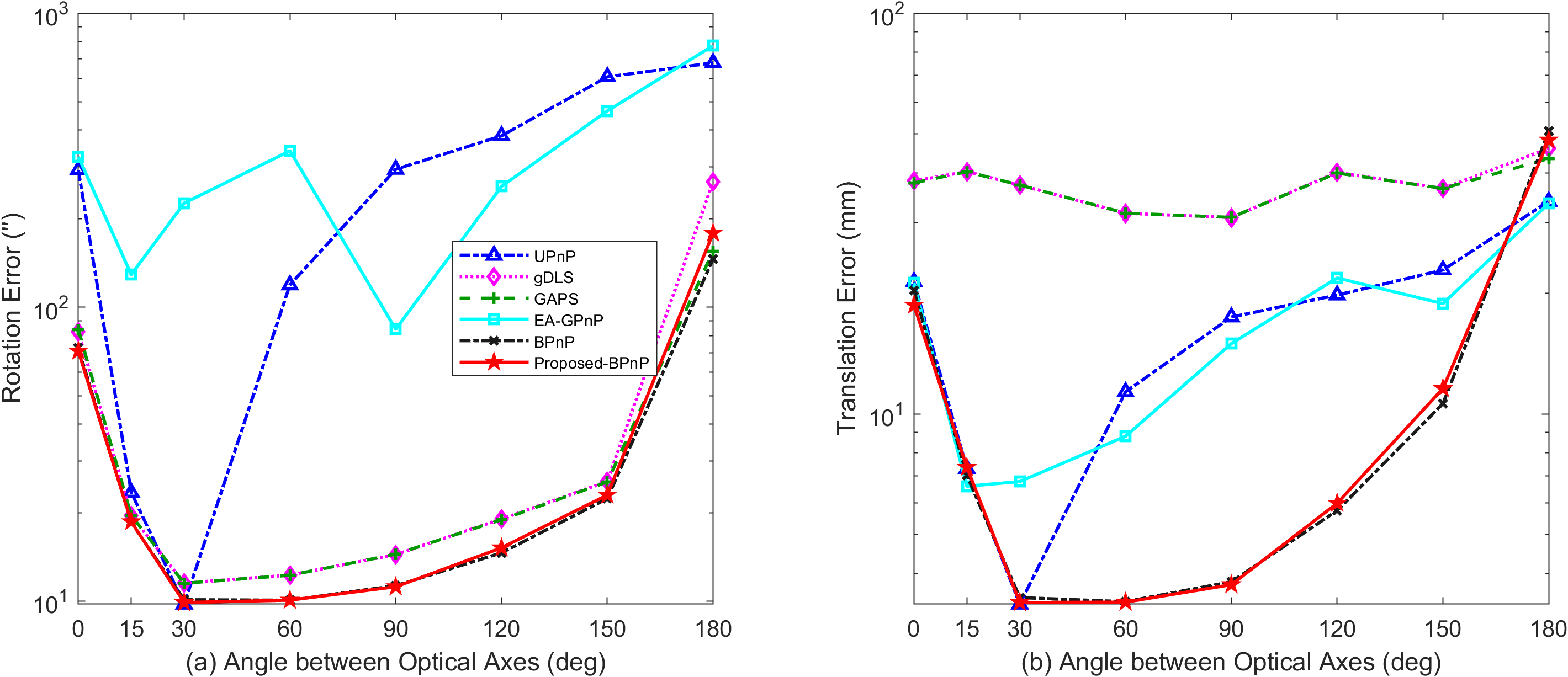}
\caption{Influence of the angle between the two optical axes: (a) rotation;
(b) translation.}
\label{fig:mountangle}
\end{figure*}

Fig.~\ref{fig:mountangle} shows the influence of the mounting angle between the two optical axes, which varies from $0^{\circ}$ to $180^{\circ}$. 
The best accuracy is obtained for angles between $30^{\circ}$ and $90^{\circ}$, where the rotation RMSE of the proposed solution stays around $10''$ and the translation RMSE around 3.4~mm; when the two axes are nearly parallel ($0^{\circ}$) or opposite ($180^{\circ}$), the observation geometry degenerates and the errors of all methods grow by roughly an order of magnitude. 
The $60^{\circ}$ angle adopted in the default configuration therefore lies in the optimal range.

\subsubsection{Statistical Distribution and Efficiency}
\begin{figure*}[!tb]
\centering
\includegraphics[width=0.44\textwidth]{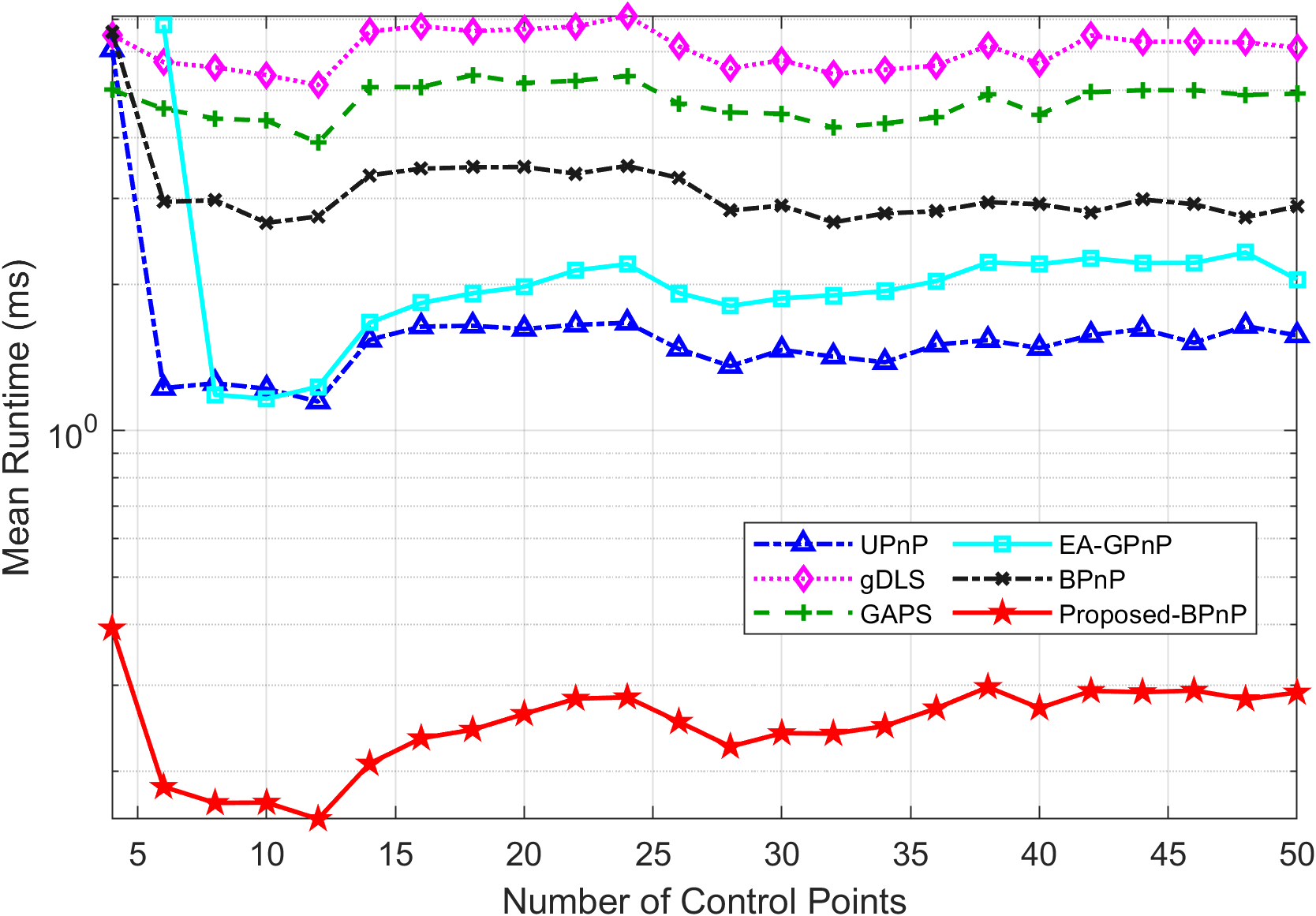}
\caption{Average runtime versus the number of control points (multi-camera
group).}
\label{fig:binoruntime}
\end{figure*}

\begin{figure*}[!tb]
\centering
\includegraphics[width=0.8\textwidth]{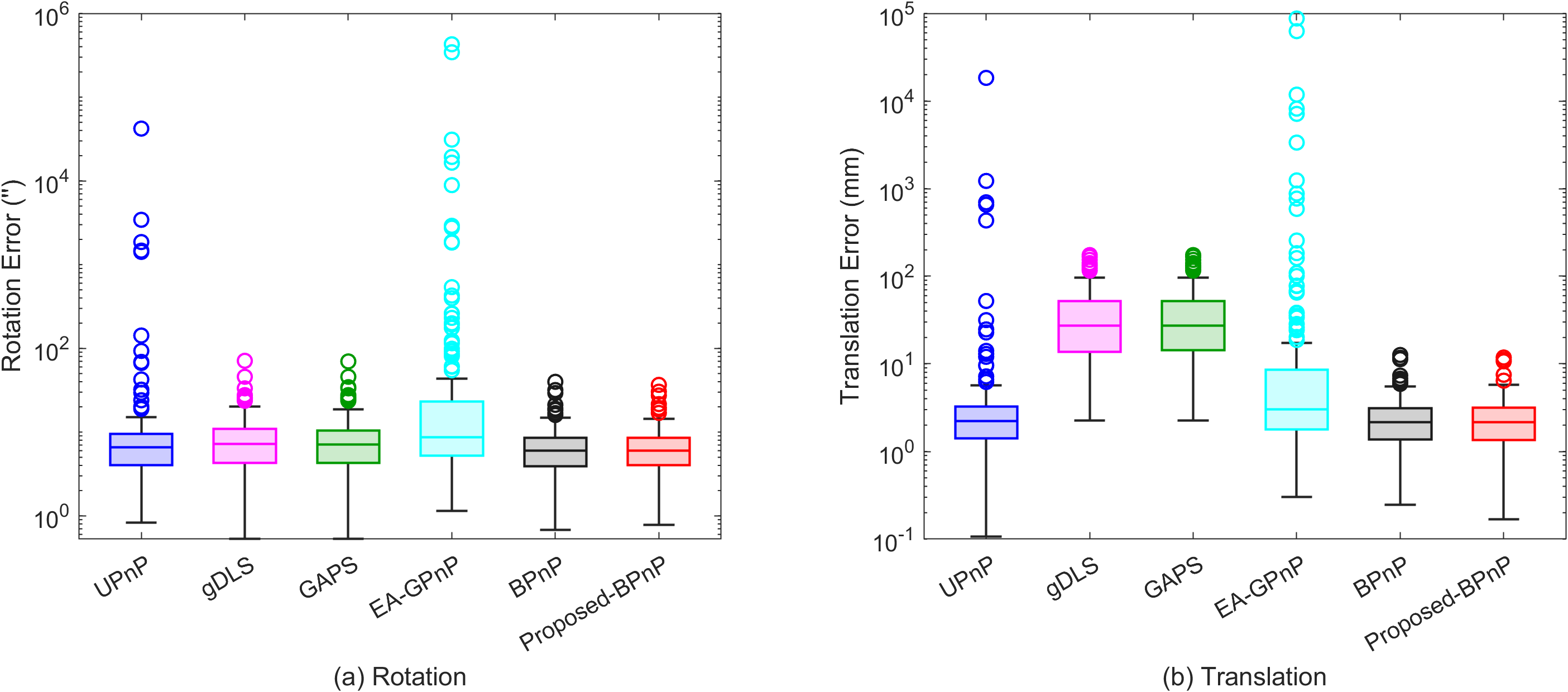}
\caption{Box plots of the rotation and translation errors at the 0.5-pixel
noise level: (a) rotation; (b) translation.}
\label{fig:binobox}
\end{figure*}

Table~\ref{tab:binostats} summarizes the statistics at the 0.5-pixel noise
level (5 control points), which matches the reprojection error observed in
the field experiment (0.50--0.57 pixel). For the rotation, the proposed
two-pass linear solution matches the LM-refined solution almost exactly
(RMSE $10.58''$ versus $10.49''$) and both outperform gDLS/GAPS (about
$13.2''$), while UPnP attains a lower median ($6.7''$) but suffers from
frequent divergence (filtered RMSE above $100''$). The advantage is more
pronounced for the translation: the proposed linear solution attains a
translation RMSE of 3.91~mm (median 2.19~mm), essentially identical to the
LM-refined solution (3.86~mm) and about one order of magnitude better than
the gDLS/GAPS solvers (47~mm RMSE, median 23~mm); UPnP reaches a low
translation median (2.25~mm) when it converges but diverges frequently.
EA-GPnP achieves a good translation median (3.04~mm) when it converges but
suffers from divergence. The SDP-based gOp, included for completeness,
diverges in 32\% of the trials and its filtered rotation RMSE exceeds
$1000''$, although its translation median (3.18~mm) is fair when it
converges; a runtime of about 170~ms further precludes real-time use. In terms of efficiency, the proposed linear solution comprises two deterministic linear passes and takes only about 0.27~ms per solution, about $4\times$ faster than the fastest generalized solver UPnP (1.15~ms) and more than an order of magnitude faster than gDLS/GAPS (4.5--5.5~ms) and the LM-refined solution (4.4~ms); 
the average runtime versus the number of control points is reported in Fig.~\ref{fig:binoruntime}. 
Hence, under micro-motion conditions, the proposed differential formulation dominates the generalized absolute-pose solvers in accuracy, robustness, and efficiency simultaneously, which is consistent with the theoretical analysis in Section~\ref{sec:theory}.
Fig.~\ref{fig:binobox} further depicts the error distributions at the 0.5-pixel noise level. The boxes and whiskers of the proposed differential solution are the lowest without heavy tails, with evident outliers for UPnP and EA-GPnP.

\begin{table}[htbp]
\setlength{\tabcolsep}{2.0pt}
\centering
\caption{Rotation and Translation Error Statistics at the 0.5-Pixel Noise Level (5 Control Points, $\pm 30'$ Rotation, $\pm 1$~mm Translation)}
\label{tab:binostats}
\footnotesize
\setlength{\tabcolsep}{2.2pt}
\begin{tabular}{lccccc}
\toprule
Method & \makecell*[c]{Rot.\\RMSE/$''$} & \makecell*[c]{Rot.\\median/$''$}
       & \makecell*[c]{Trans.\\RMSE/mm} & \makecell*[c]{Trans.\\med./mm}
       & Time/ms\\
\midrule
UPnP    & 101.29  & 6.73   & ---   & 2.25 & 1.15\\
gDLS    & 13.18   & 7.01   & 47.45 & 22.81& 5.54\\
GAPS    & 13.15   & 7.01   & 47.47 & 22.85& 4.52\\
EA-GPnP & 211.07  & 8.42   & ---   & 3.04 & 3.61\\
BPnP    & 10.49   & 6.57   & 3.86  & 2.21 & 4.40\\
gOp     & 1396.21 & 577.14 & 46.90 & 3.18 & 172.8\\
Proposed-bino (linear) & \textbf{10.58} & \textbf{6.45} & \textbf{3.91}
 & \textbf{2.19} & \textbf{0.27}\\
\bottomrule
\end{tabular}
\end{table}

\subsection{Real-World Validation (Summary)}
Beyond the synthetic studies, the proposed solvers were validated on a laboratory dual-camera platform (controlled micro-rotations; the differential solutions remain consistent with the simulation-predicted accuracy across varying numbers of control points) and in a long-term binocular deployment on the Qipanzhou Bridge, where the differential method exhibits visibly smaller environment-induced disturbances and better stability than the PnP baselines. 
Complete setups, calibration tables, and result figures are provided in Appendix~E of the supplemental material (Figs.~S1--S5,
Tables~S1--S2).

\subsection{Validation of the Bias-Eliminated Estimator}
\label{sec:bevalid}
To verify Theorem~\ref{thm:consistency}, Monte Carlo simulations were performed under the configuration of Section~\ref{sec:simmono} ($f=100{,}000$ pixels, $3840\times 2160$ image, depths 50--100~m) with a micro-motion of $2'$ rotation and 0.5~mm translation, so that the linearization residual is negligible and the errors-in-variables effect is isolated.
The interaction matrix is constructed from the measured image points at $t_0$, and the empirical bias is measured by a control-variate scheme --- the mean difference from the noise-free-regressor solution under identical noise realizations --- with 2500 trials per configuration.

Fig.~\ref{fig:bebias} summarizes the results. The LS bias remains at a constant plateau ($\approx 0.097$~mm at $\sigma=1$ pixel) regardless of the number of points, whereas the BE bias decays approximately as $1/n$ --- faster than the $1/\sqrt{n}$ statistical fluctuation --- confirming the consistency statement of Theorem~\ref{thm:consistency}; the residual-based
estimate of \eqref{eq:sigmahat} is indistinguishable from using the true noise level. The bias grows exactly quadratically with the pixel noise, matching \eqref{eq:moments}, and is confined to the axial translation component, matching the depth-rate structure of \eqref{eq:dM}. 
The practical implication is that in single-epoch solutions with few points the bias is buried under random error, but in network or long-term averaging regimes, where random errors average out, the LS bias persists as a systematic floor while the BE estimator removes it at no extra cost.

\FloatBarrier

\begin{strip}
\centering
\includegraphics[width=0.85\textwidth]{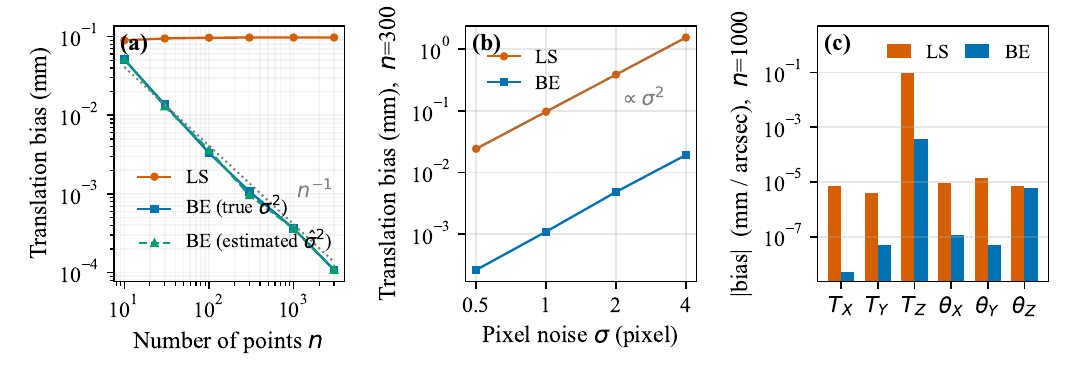}
\captionof{figure}{Empirical biases of the least-squares (LS) and bias-eliminated (BE) estimators. Panel (a) shows the translation bias as the number of points $n$ increases at $\sigma=1$ pixel. The LS bias remains approximately constant, whereas the BE bias decays as $1/n$; the curves obtained using the true and residual-estimated values of $\sigma^2$ coincide. Panel (b)
shows that the LS bias follows the $\sigma^2$ dependence predicted by \eqref{eq:moments} when $n=300$. Panel (c) shows that, when $n=1000$, the LS bias is concentrated in the axial translation $T_Z$ and is removed by the BE correction.}
\label{fig:bebias}
\end{strip}

\section{Conclusion}
This work reformulates 6-DOF platform-motion estimation as a differential 3D-2D problem rather than the difference between two independently estimated absolute poses. The resulting formulation admits an efficient closed-form linear solution based on inter-frame image displacements. 
More importantly, it exposes the distinct roles of the two components of extrinsic calibration error: translational errors are eliminated exactly, whereas rotational errors induce only bounded perturbations within the admissible motion range. 
The accompanying observability and approximation analyses clarify when the differential model is solvable and when its small-motion assumptions remain valid.

The proposed solvers achieve state-of-the-art accuracy and the fastest runtime among the evaluated monocular and multi-camera methods, while requiring substantially weaker extrinsic-calibration accuracy than conventional PnP pipelines. These properties make the method particularly suitable for high-precision micro-motion measurement and long-term structural monitoring.

\FloatBarrier
\bibliographystyle{IEEEtran}
\bibliography{IEEEabrv,./ref.bib}

\end{document}